\documentclass{article} %
\usepackage{iclr2027_conference,times}

\usepackage{amsmath,amsfonts,bm}

\def\eqref#1{equation~\ref{#1}}

\def\floor#1{\lfloor #1 \rfloor}
\def\1{\bm{1}}

\DeclareMathAlphabet{\mathsfit}{\encodingdefault}{\sfdefault}{m}{sl}
\SetMathAlphabet{\mathsfit}{bold}{\encodingdefault}{\sfdefault}{bx}{n}

\usepackage{inconsolata}

\usepackage{amsthm}
\newtheorem{theorem}{Theorem}
\newtheorem{lemma}{Lemma}
\newtheorem{corollary}{Corollary}
\newtheorem{proposition}{Proposition}
\theoremstyle{definition} 
\newtheorem{definition}{Definition}

\makeatletter
\def\thm@space@setup{%
  \thm@preskip=3pt plus 1pt minus 1pt   %
  \thm@postskip=0pt            %
}
\makeatother

\usepackage{amssymb}
\usepackage{paralist}
\usepackage{thm-restate}

\usepackage{booktabs}

\usepackage{mathtools}

\DeclarePairedDelimiter{\Abs}{\lVert}{\rVert}

\usepackage{tabularx}
\usepackage{siunitx}
\newcommand{\rng}[2]{{\footnotesize\color{gray}[#1--#2]}}

\newcommand\D{\mathbb D}

\newcommand\Z{\mathbb Z}

\newcommand\overflow{\bot}

\newcommand\AC{\mathsf{AC}}
\newcommand\ACC{\mathsf{ACC}}
\newcommand\TC{\mathsf{TC}}
\newcommand\NC{\mathsf{NC}}

\newcommand\Rtrv{\mathsf{Rtrv}}
\newcommand\Aprd{\mathsf{Aprd}}
\newcommand\Sol{\mathsf{Sol}}
\newcommand\Fin{\mathsf{Fin}}

\newcommand\AHAT{\mathsf{AHAT}}
\newcommand\SMAT{\mathsf{SMAT}}
\newcommand\GDN{\mathsf{GDN}}

\newcommand\modgadget{\mathsf{GM}}

\newcommand\FO{\mathsf{FO}}
\newcommand\MOD{\mathsf{MOD}}
\newcommand\LTL{\mathsf{LTL}}
\newcommand{\LTLpast}{\LTL[\mathsf{past}]}
\newcommand\MSO{\mathsf{MSO}}

\let\defeq\triangleq

\usepackage{xcolor}
\definecolor{darkblue}{RGB}{0, 51, 153}
\usepackage[colorlinks=true, linkcolor=darkblue, citecolor=darkblue, urlcolor=darkblue, filecolor=darkblue]{hyperref}

\usepackage{url}
\usepackage{graphicx}

\usepackage[capitalize,noabbrev]{cleveref}
\crefname{section}{\S\!}{\S\S\!}
\crefname{appendix}{\S\!}{\S\S\!}
\crefname{table}{table}{tables}
\Crefname{table}{Table}{Tables}

\usepackage{tikz}
\definecolor{stbase}{HTML}{F28F3B}  %

\newcommand{\sbhl}[1]{{%
  \setlength{\fboxsep}{1.5pt}\setlength{\fboxrule}{0.5pt}%
  \smash{\fcolorbox{stbase!80}{stbase!20}{#1}}}}

\usepackage{listings}
\usepackage{xcolor}

\definecolor{sbkey}{HTML}{35618C}
\definecolor{sbnum}{HTML}{A03050}
\definecolor{sbrule}{HTML}{CCCCCC}

\lstdefinestyle{sbcode}{
  basicstyle=\ttfamily\scriptsize,
  keywordstyle=\color{sbkey}\bfseries,
  emphstyle=\color{sbnum},
  frame=single, rulecolor=\color{sbrule}, framesep=6pt,
  xleftmargin=6pt, xrightmargin=6pt,
  columns=fullflexible, keepspaces=true,
  showstringspaces=false, aboveskip=0pt, belowskip=0pt,
  morekeywords={int},
  emph={9,10,24,37,58}
}

\lstdefinestyle{sbprose}{
  basicstyle=\ttfamily\scriptsize,
  frame=single, rulecolor=\color{sbrule}, framesep=6pt,
  xleftmargin=6pt, xrightmargin=6pt,
  columns=fullflexible, keepspaces=true,
  breaklines=true, aboveskip=0pt, belowskip=0pt,
  literate={pink}{{\color{sbnum}pink}}4
           {orange}{{\color{sbnum}orange}}6
           {olive}{{\color{sbnum}olive}}5
           {green}{{\color{sbnum}green}}5
           {blue}{{\color{sbnum}blue}}4
}

\newcommand{\changed}[1]{\textcolor{red}{#1}}

\title{Measuring the Fine Structure of State-Tracking in Language Models with \textbf{StateBench}}
\title{A Fine-Grained Hierarchy and Benchmark for State Tracking in Language Models}

\author{Jackson Petty$^{\rtimes,\oplus}$ \quad William Merrill$^{\rtimes,\otimes}$ \quad Ashish Sabharwal$^{\rtimes}$ \\
$^{\rtimes}$Allen Institute for AI \quad $^{\oplus}$New York University \quad
$^{\otimes}$Toyota Technological Institute at Chicago\\
\texttt{research@jacksonpetty.org}
}

\iclrfinalcopy %
\begin{document}

\maketitle

\fancyhead{}

\begin{abstract}
Theoretical limitations of transformers on \emph{state tracking}---the ability to track information about entities or variables over long contexts---have motivated new recurrent language modeling (LM) architectures such as Gated DeltaNet (GDN). These architectures have, in particular, been benchmarked against the $S_5$ word problem, a theoretically hard state-tracking task that captures abilities such as code evaluation under arbitrary variable assignment. It, however, remains open whether other, simpler forms of state tracking might both be more relevant in practice and already explain observed differences between these architectures. We propose evaluating state tracking using a hierarchy of word problems with four increasing levels of difficulty: \emph{r-trivial}, corresponding to immutable assignment to variables in programming; \emph{aperiodic}, where assignments are mutable; \emph{soluble}, where assignments are self-referential; and \emph{general} finite state tracking. Theoretically, we show that the class of problems transformers are thought to length-generalize on does not even contain all aperiodic state tracking. On the other hand, we prove that GDN with positive eigenvalues expresses \emph{all} aperiodic problems, and the regular languages recognized by certain transformers are precisely all soluble ones, thus providing a fine-grained picture of state tracking capabilities of various architectures. Taking these ideas to practice, we construct \textbf{StateBench}, a 4-tier long-horizon state-tracking benchmark. Our systematic direct training experiments across architectures largely align with theoretical expectation, albeit with interesting observations at very large scale length-generalization. Additionally, unlike prior work, we also evaluate pretrained models---both open hybrid and closed frontier ones---on StateBench, finding that none of them robustly solves even soluble state tracking, suggesting a direction for future improvement.
\end{abstract}

\section{Introduction}
Tracking the state of the world across a long context is necessary for understanding discourse, documents, and code \citep{kamp-1981-theory,kim-schuster-2023-entity,hsieh-2024-ruler-a}. Prior work has examined the difficulty of state-tracking tasks by viewing them as equivalent to word problems for finite monoids, and relating the algebraic properties of these structures to their computability by various neural architectures \citep{liu2023transformers,merrill-2024-illusion}.
While much focus has been given to the hardest class of state-tracking problems, such as the $S_5$ word problem, less attention has been given to intermediate levels of problem difficulty, which may prove more representative of the kinds of real-world state-tracking tasks language models are expected to solve and might already explain observed performance differences across architectures.

To better understand finegrained state-tracking differences between architectures, we propose evaluating state tracking capabilities across four hierarchical levels naturally motivated by the classical Krohn--Rhodes decomposition for finite monoids \citep{krohn-1965-algebraic}:
\emph{general finite}, involving arbitrary transpositions of variables; \emph{soluble}, involving variable assignment by self-reference; \emph{aperiodic}, involving the reassignment of variables to new values; and \emph{r-trivial}, involving constant registers which can only be written to once.
We then evaluate the theoretical expressivity and empirical performance of various language model architectures across levels of this hierarchy. 

We first theoretically explore the ability of Transformers \citep{vaswani-2017-attention} and Gated DeltaNet (GDN; \citealt{yang-2024-gated}) models to represent state tracking problems across this hierarchy and the role of particular hyperparameters.
Complementing existing results relating low-precision softmax transformers to r-trivial and aperiodic state tracking \citep{li-2026-characterizing,yang-2025-knee-deep}, we show that the state tracking problems expressible by log-precision averaging hard-attention transformers, a well-studied theoretical model, are exactly the soluble ones.
Next, whereas GDN with an extended eigenvalue spectrum can represent all state tracking tasks, we characterize the state tracking capabilities of GDN with positive eigenvalues as containing all
aperiodic monoids and even some periodic (soluble) ones.
Synthesizing prior work and our new results reveals finegrained distinctions between the theoretical state tracking capabilities of transformers and GDN, as well as the role of precision and parameterization details on the state tracking expectations for each architecture.

Motivated by these theoretical results,
we introduce StateBench, a long-context evaluation benchmark with state-tracking problems at varying levels of complexity and supporting evaluation of very large scale length-generalization (up to 64K state changes). We evaluate models trained directly on StateBench data, showing that models containing recurrent modules length-generalize fully on r-trivial, aperiodic, and soluble data and show limited generalization on general finite data, while transformers fail to generalize on r-trivial data and fail to learn even in-distribution general finite data.
Surprisingly given prior work \citep{grazzi2025unlocking}, we find that recurrent models with non-negative eigenvalues succeed on StateBench's soluble state tracking tasks, which we explore theoretically in \Cref{thm:gdn-equals-aperiodic}.
On open-source pretrained models including two Olmo 3 7B variants, we find that all architectures do well on r-trivial data, degrade meaningfully on aperiodic data, and fail completely on general finite data. We find that a variety of frontier proprietary Gemini, Claude, and GPT LLMs can solve r-trivial and aperiodic tasks well but exhibit strong degradation on soluble and general finite state tracking tasks as a function of sequence length, dropping to near-zero performance on the hardest problems in StateBench. Lastly, we contextualize these results by connecting monoid periodicity to the sensitivity of boolean functions, showing that proper understanding of many tasks involving real-world source code requires state-tracking capabilities beyond the expressivity and learned capacity of current language models.

\section{A Hierarchy of Finite State Tracking}

\begin{figure}[t]
\centering
\definecolor{stbase}{HTML}{F28F3B}
\begin{tikzpicture}[
  clabel/.style = {anchor=north west, align=left, inner sep=0pt,
                   font=\sffamily\scriptsize, text=stbase!65!black},
  rlabel/.style = {anchor=east, font=\sffamily\scriptsize, text=stbase!65!black},
  task/.style   = {align=center, font=\small\itshape},
  arch/.style   = {align=center, font=\small\sffamily},
  code/.style   = {font=\small\ttfamily},
  circ/.style   = {anchor=east, font=\small, text=stbase!65!black},
]

\def\colw{3.0}      %
\def\tailstep{0.2}  %
\def\rowtask{0}     %
\def\rowcode{-0.65}
\def\rowarch{-1.3}
\def\floor{-1.65}   %
\def\shelltop{1.2}  %
\def\tail{0.55}     %
\def\corner{4pt}    %

\newcommand{\shell}[3]{%
  \draw[line width=0.9pt, fill=#2, draw=#3, rounded corners=\corner]
    (0,\floor) -- (0,\shelltop)
    -- ({#1*\colw},\shelltop)
    [sharp corners] -- ({#1*\colw},\floor) -- cycle;}
\newcommand{\circuit}[4]{%
  \path[fill=#2, rounded corners=\corner]
    (0,\floor) -- (0,{\floor-\tail-(#1-2)*\tailstep})
    -- ({#1*\colw},{\floor-\tail-(#1-2)*\tailstep})
    [sharp corners] -- ({#1*\colw},\floor) -- cycle;
  \draw[#4, line width=0.7pt, draw=#3, rounded corners=\corner]
    (0,\floor) -- (0,{\floor-\tail-(#1-2)*\tailstep})
    -- ({#1*\colw},{\floor-\tail-(#1-2)*\tailstep}) -- ({#1*\colw},\floor);}
\newcommand{\classlabel}[3]{%
  \node[clabel] at ({(#1-1)*\colw+0.15},{\shelltop-0.15})
    {\textbf{#2}\\#3};}
\newcommand{\splitarch}[5][]{%
  \draw[#2, line width=1pt, draw=stbase!80]
    ({(#3-0.5)*\colw},{\rowarch+0.3}) -- ({(#3-0.5)*\colw},{\rowarch-0.3});
  \node[arch, #1] at ({(#3-0.75)*\colw},\rowarch) {#4};
  \node[arch] at ({(#3-0.25)*\colw},\rowarch) {#5};}

\circuit{4}{stbase!5}{stbase!80}{solid}
\circuit{3}{stbase!10}{stbase!80}{dashed}
\circuit{2}{stbase!20}{stbase!80}{solid}
\shell{4}{stbase!5}{stbase!80}
\shell{3}{stbase!10}{stbase!80}
\shell{2}{stbase!20}{stbase!80}
\shell{1}{stbase!30}{stbase!80}

\foreach \k/\class in {2/{\AC^0}, 3/{\TC^0}, 4/{\NC^1}}
  \node[circ] at ({\k*\colw-0.25},{\floor-0.3}) {$\class$};

\classlabel{1}{R-TRIVIAL}{$\LTLpast$}
\classlabel{2}{APERIODIC}{$\LTL$ or $\FO$}
\classlabel{3}{SOLUBLE}{$\FO + \MOD$}
\classlabel{4}{GENERAL FINITE}{$\MSO$}

\foreach \y/\row in {\rowtask/tasks, \rowcode/data, \rowarch/arch}
  \node[rlabel] at (-0.15,\y) {\row};

\node[task] at (0.5*\colw,\rowtask) {Immutable\\assignment};
\node[task] at (1.5*\colw,\rowtask) {Assignment\\by value};
\node[task] at (2.5*\colw,\rowtask) {Assignment by\\self-reference};
\node[task] at (3.5*\colw,\rowtask) {Assignment by\\reference};

\node[code] at (0.5*\colw,\rowcode) {let x = 4};
\node[code] at (1.5*\colw,\rowcode) {var x = 4};
\node[code] at (2.5*\colw,\rowcode) {var x = x + 5};
\node[code] at (3.5*\colw,\rowcode) {var x = y};

\node[arch] at (0.5*\colw,\rowarch) {$\mathsf{SMAT}_f$};
\splitarch{solid}{2}{$\mathsf{SMAT}_m$}{}
\splitarch{dashed}{3}{$\mathsf{GDN}^{+}_f$}{AHAT$_\ell$}
\node[arch] at (3.5*\colw,\rowarch) {$\mathsf{GDN}^{\pm}_f$};
\end{tikzpicture}
\caption{Placement of various language model architectures within the \emph{hierarchy of state-tracking} (monoid word problem) complexity, along with examples of the kinds of tasks that fall into these categories.
Areas below the long horizontal line indicate circuit complexity classes that, when restricted to regular languages, are equal to the monoid classes above them. The two log-precision architectures extend into $\TC^0$. Dashed lines represent conjectured strict boundaries between classes.}
\label{fig:hierarchy}
\end{figure}
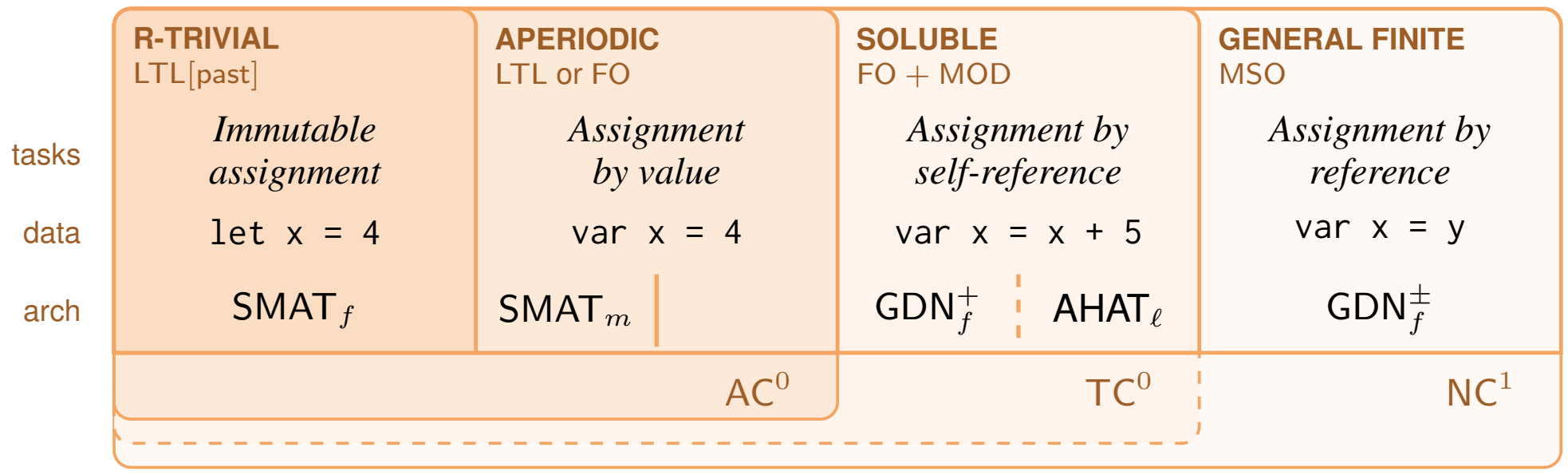

To analyze the difficulty of state-tracking problems and connect their solvability to neural architectures, we construct a formal model of them based on an algebraic object known as a \emph{(finite) monoid}, defined as a triple $(M,\circ,1)$ of some (finite) set $M$, a composition operation $\circ$, and a neutral identity element $1$. Prior work has noted that monoids are a natural model of state tracking by means of their composition operator~\citep[][\emph{inter alia}]{liu2023transformers,merrill-2024-illusion,nowak-2026-algebraic,nowak-2026-compositional}. A \emph{word} $w$ over $M$ is a string $w_1 w_2 \ldots w_n$ of elements of $M$ concatenated together, representing the composition $w_1 \circ w_2 \circ \cdots \circ w_n$. Intuitively, $w$ represents a sequence of updates to a system: given some initial state $s_0$, observing $w_1$ moves the system to state $s_1$; observing $w_2$ moves it to $s_2$; and so forth. Since $M$ is closed under composition, the word $w = w_1w_2\dots w_n$ is itself an element of $M$; computing exactly which element is known as the \emph{word problem}\footnote{For brevity, we will often omit writing the operator $\circ$ explicitly, treat composition as concatenation, and refer to a monoid $(M,\circ,1)$ simply by its set $M$. As a canonical example, consider the integers $(\Z,+,0)$ with addition; the word problem on $\Z$ is identifying $a_1 + a_2 + \dots + a_k$ for any $a_1,\dotsc,a_k \in \Z$.} on $M$.

We will formalize different levels of complexity for state tracking in terms of word problems for different types of monoids.
A useful source of intuition for this will be the different kinds of variable state manipulation that are possible in programming.
We can distinguish between four kinds of state updates: (i) \emph{immutable writes}, such as defining a constant value like \texttt{let x = 4}; (ii) \emph{mutable writes}, such a (re)defining a variable value like \texttt{var x = 4; ... x = 2}; (iii) \emph{self-referential writes}, such as incrementing a variable like \texttt{var x = 4; x += 3}; and (iv) \emph{referential writes}, such as swapping two variables like \texttt{var x = 4; var y = 2; (x,y) = (y,x)}.

By modeling state-tracking as a word problem on a finite monoid, we can relate these various kinds of state update operations to the algebraic structure of the monoid and use this to understand the computational difficulty of tracking the state-space---calculating the value of the system after observing an arbitrary sequence of updates. The well-known eponymous theorem of \citet{krohn-1965-algebraic} establishes that every finite monoid~$M$ can be decomposed into the iterated wreath product, denoted $M_1 \wr M_2 \wr \cdots \wr M_k$, of finitely-many irreducible factors, loosely analogous to a (non-unique) prime factorization.\footnote{We defer formal definitions of the algebraic concepts related to monoids and Krohn--Rhodes decomposition to \Cref{sec:formal-defs}, focusing here on developing the intuition behind the state tracking hierarchy.}
Each irreducible factor $M_i$ in this decomposition is either the two-element semilattice \smash{$U_1 \defeq \{1,0\}$} where $0x = 0$, corresponding to \emph{immutable writes}; the ``flip-flop'' monoid \smash{$U_2 \defeq \{1,a,b\}$} where $xy =  y$ for $x,y \in \{a,b\}$, corresponding to \emph{mutable writes} by value; the commutative group $\Z_p$ of integers modulo $p$ for some prime $p$, corresponding to \emph{self-referential writes}; or a non-commutative finite simple group, corresponding to \emph{referential writes}.

Different kinds of state tracking can be formalized based on the kinds of finite monoids they represent. A finite monoid $M$ is called
  (i) \emph{r-trivial} if for all $x, y \in M$, $xM = yM$ implies $x = y$ \citep{green-1951-structure};
  (ii)~\emph{aperiodic} if for every $x \in M$ there is some $n \geq 1$ with
  $x^n = x^{n+1}$ \citep{schutzenberger-1965-finite}; and
  (iii)~\emph{soluble} if every subgroup of $M$ is soluble \citep{barrington-1988-finite}.
For our purposes, however, a more useful and equivalent characterization of these types of monoids is terms of the kinds of core operations that appear in their Krohn--Rhodes (KR) decomposition:

\begin{proposition}[Monoid types, informal; cf.~\cref{sec:formal-defs} for a formal statement] \label{def:monoid-types}
  A finite monoid $M$ with KR decomposition $M_1 \wr M_2 \wr \cdots \wr M_k$ is
  \emph{(i)~r-trivial} if every $M_i$ is $U_1$;
  \emph{(ii)~aperiodic} if every $M_i$ is $U_2$;\footnote{$U_1$ is a submonoid of $U_2$, obtained by setting $a = b = 0$. $U_2$ thus includes $U_1$ in a natural way.}
  and
  \emph{(iii)~soluble} if every $M_i$ is either $U_2$ or $\Z_p$ for some prime $p$.
\end{proposition}

As illustrated in \Cref{fig:hierarchy}, these classes form a nested hierarchy:
\sbhl{$\Rtrv \, \subsetneq \, \Aprd \, \subsetneq \, \Sol \, \subsetneq \, \Fin$},
where $\Rtrv$ is the class of word problem languages\footnote{For a monoid $M$, $w$ is in the word problem language \smash{$L(M)$} if the product of $w$ evaluates to $1$.} whose monoid is r-trivial, $\Aprd$ is the class for aperiodic monoids, $\Sol$ is the class for soluble, and $\Fin$ contains all word problem languages on finite monoids including $S_5$.
Thus, $\Rtrv$ can be understood as a generalized class of all state tracking involving only immutable writes, $\Aprd$ as a natural class of state tracking problems involving only writes ``by value'', $\Sol$ as a harder class of state tracking where interaction between variables is limited, and $\Fin$ as the class of unconstrained finite state tracking problems.

\noindent \textbf{Correspondence to Formal Language, Logic, and Circuit Complexity Classes.}
In addition to having intuitive interpretations in terms of variable assignments, these levels within the state tracking hierarchy coincide with other notable classes of problems.
From a formal language theory perspective, $\Fin$ is exactly the class of regular languages, and $\Aprd$ is exactly the class of star-free languages \citep{schutzenberger-1965-finite}.
From a logical perspective, $\Fin = \MSO$, the languages definable in monadic second-order logic \citep{Buchi1990}, whereas $\Aprd = \FO[<] = \LTL$: it corresponds to the languages definable in first-order logic~\citep{mcnaughton1971counter} or, equivalently, linear temporal logic~\citep{Kamp1968}.
$\Sol = \FO[<] + \MOD$, the languages definable in first-order logic augmented with modular predicates \citep{tesson-2007-logic}, and $\Rtrv = \LTL[\mathsf{past}]$, the languages definable in the restricted past fragment of $\FO[<]$ or $\LTL$ \citep{li-2026-characterizing,jerad-2025-unique}.

Finally, from a circuit complexity perspective, $\Aprd$ corresponds to the regular languages in $\AC^0$ \citep[Theorem 3]{mixBarrington-1992-regular}, $\Fin$ to the regular languages in $\NC^1$ (all of them), and $\Sol$ to the regular languages in $\ACC^0$ as well as those in $\TC^0$, assuming $\TC^0 \neq \NC^1$ \citep[Theorem 8]{mixBarrington-1992-regular}.
The convergence of these classes with other areas suggests the hierarchy reflects fundamental distinctions in state tracking capabilities relevant across computational models.

\section{Expressivity of Architectures for State Tracking}

We now explore the state tracking abilities of various transformer and recurrent architectures, in relation to our state tracking hierarchy.
Our goal is to characterize models based on whether or not they can express a length-general solution to the word problem for the particular class.

\noindent \textbf{Transformers and GDN.}
We adopt the formal model of \citet{merrill-2026-why} and \citet{merrill2025theory}; see \Cref{app:transformer-model} for formal definitions. 
Each architecture uses a stack of $d$ layers, interleaving tokenwise feedforward sublayers with sequence-mixing sublayers that are defined as follows:

\begin{compactitem}\vspace{-0.7ex}
    \item Softmax attention \textbf{(SMAT)}: the standard causally-masked attention of \citet{vaswani-2017-attention}, where token $i$ outputs $\sum_{j \leq i} \mathrm{softmax}_j(\mathbf q_i^\top \mathbf k_j)\, \mathbf v_j$.
    
    \item Averaging hard attention \textbf{(AHAT)}: an idealization in which token $i$ attends uniformly to tied positions $j \leq i$ that maximize $\mathbf q_i^\top \mathbf k_j$ and averages their values \citep{perez-2021-attention}.
    
    \item Gated DeltaNet \textbf{(GDN)} \citep{yang-2024-gated}: a linear recurrence over a matrix-valued state,
    \begin{equation}
        \mathbf S_t = \alpha_t \big(\mathbf I - \beta_t \mathbf k_t \mathbf k_t^\top\big) \mathbf S_{t-1} + \beta_t \mathbf k_t \mathbf v_t^\top, \qquad \mathbf o_t = \mathbf S_t^\top \mathbf q_t,
    \end{equation}
    with gate $\alpha_t \in [0,1]$ and write strength $\beta_t$. The eigenvalues of the transition $\mathbf I - \beta_t \mathbf k_t \mathbf k_t^\top$ are $1$ and $1 - \beta_t$. Following \citet{grazzi2025unlocking}, we write \smash{$\GDN^+$} when $\beta_t \in [0,1]$ (nonnegative eigenvalues) and \smash{$\GDN^\pm$} when $\beta_t \in [0,2]$ (eigenvalues in $[-1,1]$).
\end{compactitem}

\noindent \textbf{Arithmetic Precision.}
The expressive power of these models is well-known to depend on the numerical precision of the network's activations.
Following \citet{merrill-2026-why}, we distinguish \emph{fixed} precision ($f$; a fixed number of bits w.r.t.\ input length $n$), \emph{log} precision ($\ell$; $O(\log n)$ bits, enough to represent positions and uniform averages over the context), and \emph{mixed} precision ($m$; fixed precision everywhere except inside attention aggregation, as in \citealt{yang-2025-knee-deep}). Our fixed-precision analysis uses a \emph{fixed-precision, fixed-point overflow datatype}, $\D_{p,d}$, defined in \Cref{sec:fixed-precision}. Informally, it represents numbers in the form $a/d$ where $d$ is a fixed positive denominator that controls the rounding behavior, and $a$ is either an integer in the range $[-2^p, 2^p]$ or a special overflow value $\overflow$. We write, e.g., \smash{$\SMAT_f$} and $\GDN_f$ for the classes of languages recognized by fixed-precision SMATs and GDNs, respectively. Throughout, we say an architecture \emph{solves} the word problem over $M$ if a single model in that architecture computes the correct output for every word over $M$.

\subsection{Known expressivity results} \label{sec:known-results}

\noindent \textbf{Transformers.}
We first review some known results relating softmax transformers (SMAT) to different levels in the state tracking hierarchy, parameterized by different levels of precision (finite $f$, mixed $m$, or log $\ell$).
First, \citet{li-2026-characterizing} and \citet{jerad-2025-unique} showed \smash{$\SMAT_f = \Rtrv$}, i.e.,
fixed-precision SMAT solves precisely the word problem over r-trivial monoids.\footnote{Technically, this characterization is sensitive to the arithmetic model: \citet{nowak-2026-compositional} show that under standard left-to-right floating-point summation, fixed-precision softmax attention can exceed $\Rtrv$ and recognize some aperiodic languages, while variants of attention give definite, r-trivial, or locally r-trivial classes. Across these arithmetic models, then, $\Rtrv \subseteq \SMAT_f \subseteq \Aprd$.}
Second, \cite{yang-2026-algebraic} showed that the language C-RASP, often used as an abstraction of transformers, is equivalent to the wreath product closure of bounded-depth Dyck languages, which is a strict subset of aperiodic. C-RASP is known to be equivalent to mixed-precision SMATs \citep{yang-2025-knee-deep}. It follows that $\SMAT_m \subset \Aprd$, i.e., only for a strict subset of aperiodic monoids $M$ can mixed-precision transformers solve the word problem over $M$.
Thus, while going from finite precision to mixed precision increases SMAT expressivity above $\Rtrv$, mixed-precision SMATs remain properly within $\Aprd$.
Third, even in the log-precision regime, both SMAT and AHAT are known to be in the circuit class uniform \smash{$\TC^0$} \citep{Merrill2023-mp}. Moreover, under the circuit complexity assumption \smash{$\TC^0 \neq \NC^1$}, the regular languages in \smash{$\TC^0$} are exactly $\Sol$ \citep[Theorem 8]{mixBarrington-1992-regular}. Thus, assuming \smash{$\TC^0 \neq \NC^1$}, we have both \smash{$\SMAT_\ell \cap \Fin \subseteq \Sol$} and \smash{$\AHAT_\ell \cap \Fin \subseteq \Sol$}.
Overall, while more precision may increase the state tracking capabilities of SMAT, they remain within $\Sol$, and it remains unknown whether they fully realize $\Sol$.

\noindent \textbf{GDN.}
Beyond transformers, we consider some partial results about GDNs. \citet{sarrof-2024-expressive} derived an upper bound of (word problems over) aperiodic monoids for
diagonal RNNs with non-negative eigenvalues.
\citet{grazzi2025unlocking} showed \smash{GDN$^+$} cannot represent certain periodic languages like \smash{$(aa)^*$},
but they assumed exact arithmetic, in contrast to our fixed-precision arithmetic model that allows rounding.
\citet{merrill-2026-why} show \smash{$\GDN^\pm_f = \Fin$}, i.e., for any finite monoid, there exists a fixed-precision GDN (with both positive and negative eigenvalues) that solves the word problem over it.\footnote{This result relies on pre-pending a fixed, input-independent prefix to the input.} It is worth noting that this captures the full gamut of finite state tracking problems, implying, in particular, that hybridization of fixed-precision GDN$^\pm$ with SMAT or AHAT does not further increase the architecture's expressivity for state tracking.

\subsection{New results leveraging Krohn--Rhodes decomposition}

We now leverage the Krohn--Rhodes view of state tracking to derive several new results placing transformer and GDN variants at various levels of the state-tracking hierarchy. These results help relate the state tracking capabilities of the two different architectures and elucidate the role of key parameters like precision and attention type for state tracking.
Proofs are deferred to \Cref{sec:proofs} throughout.

The following lemma provides the foundation for our formal results, identifying the ``building block'' monoids ($U_1, U_2$, and $\mathbb Z_p$) that various architectures can implement. Here, ``implement'' means that given a one-hot representation of a monoid element
$w_i$ in the input-layer hidden state at token $i$, there will be a subspace of the representation of token $i$ in the final layer that encodes the partial product $w_1 w_2 \ldots w_i$ as a one-hot vector.

\begin{restatable}[Implementing Basic Monoids]{lemma}{basicMonids}\label{lem:building-blocks}
    Various attention and recurrent mechanisms can solve various basic finite monoid word problems, as follows:
    \begin{compactenum}[\quad\quad a.] \vspace{-0.7ex}
        \item \label{lem:smat_U2} There exists a one layer SMAT$_f$ that implements $U_1$.

        \item \label{lem:gdn-flip-flop} There exists a one layer GDN$^+_f$ that implements $U_2$.

        \item \label{lem:ahat-flip-flop} There exists a one layer AHAT$_\ell$ that implements $U_2$.

        \item \label{lem:ahat-mod} For any prime $p$, there exists an $8$-layer AHAT$_\ell$ that implements $\Z_p$.
    \end{compactenum}
\end{restatable}

\Cref{lem:building-blocks} shows that for each basic monoid, there is a block of layers of various types of architectures that can be used as a gadget to implement that monoid.
Moreover, the iterated wreath product of $k$ basic monoids can be implemented by a depth-wise composition of the corresponding gadgets:

\begin{restatable}[Implementing Monoid Composition]{lemma}{monoidComposition} \label{lem:composition}
    Let $M$ be a finite monoid with a KR decomposition \smash{$M_1 \wr M_2 \wr \cdots \wr M_k$}. Suppose that for each $i$, an attention-based or recurrent architecture $A_i$ of depth $d_i$ can implement $M_i$ (potentially each with its own datatype). Let \smash{$d \defeq \sum_{i=1}^k d_i$}. Then the depth-$d$ composition \smash{$A_k \circ \ldots \circ A_2 \circ A_1$} of the architectures implements $M$.
\end{restatable}

Armed with these lemmas, we are ready to state our main technical findings, in which $M$ refers to a finite monoid.
Our first result shows that GDN$^+_f$ can solve \emph{all} aperiodic monoid word problems, and even some periodic (soluble) ones:

\begin{restatable}[$\GDN^+_f \supsetneq \Aprd$]{theorem}{gdnAperiodic}
\label{thm:gdn-equals-aperiodic}
    For any aperiodic $M$ and for some periodic but soluble $M$, there exists a fixed-precision GDN$^+$ that solves the word problem over $M$.
\end{restatable}

Our proof leverages the fact that aperiodic monoids can be decomposed into flip-flops, implying via \Cref{lem:building-blocks}.\ref{lem:gdn-flip-flop} that \smash{GDN$^+_f$} can solve all of $\Aprd$.
Specifically, for any aperiodic monoid that has a KR decomposition with $k$ basic monoids, our lower bound (\cref{lem:aperiodic-gdn} in \cref{sec:proofs}) constructs a fixed-precision GDN$^+$ of depth $k$.
Our result holds for (ungated) DeltaNet in addition to GDN, but we state it with the more common GDN acronym. To prove that this lower bound is not tight, we construct a fixed-precision GDN$^+$ that recognizes the periodic language $(aa)^*$:
\begin{restatable}{lemma}{gdnBeyondAperiodic} \label{lem:gdn-beyond-aperiodic}
    There exists a GDN$^+$ that recognizes $(aa)^*$ under datatype $\D_{p,25}$ for all large enough $p$.
\end{restatable}
This appears to contradict Theorem 1 of \citet{grazzi2025unlocking}, which states that no linear RNN with non-negative eigenvalues can recognize parity. The rectification is that their result assumes exact, unrounded multiplication.
In contrast, \Cref{lem:gdn-beyond-aperiodic} depends critically on rounding (specifically in the computation of $A h$ and $A h'$) to achieve periodic behavior. Moreover, we show (\Cref{lem:lrnn-beyond-aperiodic-exact} in \Cref{sec:proofs}) that general linear RNNs with non-negative eigenvalues can, in fact, recognize certain periodic languages even under exact arithmetic, and that this extends even to $(aa)^*$ if the $A_t$ matrices are parameterized as \emph{products} of matrices with non-negative eigenvalues (\Cref{cor:product-lrnn-aa-star} in \Cref{sec:proofs}), similar to DeltaProduct~\citep{siems-2025-deltaproduct}.

Next, we leverage Krohn--Rhodes in a similar way to obtain a lower bound on log-precision AHAT:\footnote{A similar result in a nonuniform transformer model was given by \citet[Theorem 2]{liu2023transformers}, though it was nontrivial to extend their result to the uniform AHAT model.}

\begin{restatable}[$\AHAT_\ell \supseteq \Sol$]{theorem}{ahatSoluble}
\label{thm:ahat-contains-soluble}
    For any soluble $M$, there exists a log-precision AHAT that solves the word problem over $M$.
\end{restatable}

As discussed in \Cref{sec:known-results}, assuming \smash{$\TC^0 \neq \NC^1$, $\AHAT_\ell \cap \Fin \subseteq \Sol$}. Since \smash{$\Sol \subseteq \Fin$}, \Cref{thm:ahat-contains-soluble} implies that, under this circuit complexity assumption, \smash{$\AHAT_\ell \cap \Fin$} is precisely \smash{$\Sol$}, i.e., the regular languages recognized by log-precision AHAT are exactly those whose underlying monoid is soluble:

\begin{corollary}[$\AHAT_\ell \cap \Fin \approx \Sol$] \label{cor:ahat-conditionally-equals-soluble}
    Assuming \smash{$\TC^0 \neq \NC^1$, $\AHAT_\ell \cap \Fin = \Sol$},
    i.e., there exists a log-precision AHAT that solves the word problem for $M$ if and only if $M$ is soluble.
\end{corollary}

A picture of architectures' relative state tracking capabilities emerges from these results.
For GDN, these capabilities are controlled by the eigenvalues, not gating: \smash{GDN$^+$} is limited to aperiodic problems while \smash{GDN$^\pm$} covers all finite monoids. On the other hand, AHAT with log precision sits in-between, solving word problems over soluble monoids.
For SMAT, precision controls expressivity, with mixed- and fixed-precision transformers corresponding to proper subclasses of aperiodic. The exact expressivity of log-precision SMAT or GDN$^+$ on state tracking problems remains open.

Another interesting open question is: \emph{Could a GDN$^+$ based hybrid model, possibly under higher precision, solve word problems over soluble monoids?}
Krohn--Rhodes theory lets us derive a conditional result. For any prime $p$, suppose we have an architecture $\modgadget^p$ (a \emph{mod gadget}) that, given a number $z \in \Z_{\geq 0}$ (represented as $1/z$ or layer-norm hash $\phi(z)$; \citealp{merrill-2023-expressive}) as input, can compute a representation of $z \bmod p$. \Cref{prop:conditional-hybrid-soluble} (cf.~\Cref{sec:proofs}) leverages KR decompositions to show that a hybrid model consisting of this gadget, along with \smash{GDN$^+_f$} and SMAT$_\ell$, covers \emph{all} soluble monoids.
It remains open whether GDN$^+$ with mixed- or log-precision is contained in aperiodic or, on the other hand, can implement the mod operation; if it's the latter, \Cref{prop:conditional-hybrid-soluble} shows that hybrid GDN models would capture all soluble monoids.

\section{The StateBench Dataset}

\begin{figure}[t]
    \centering
    \includegraphics[width=\linewidth]{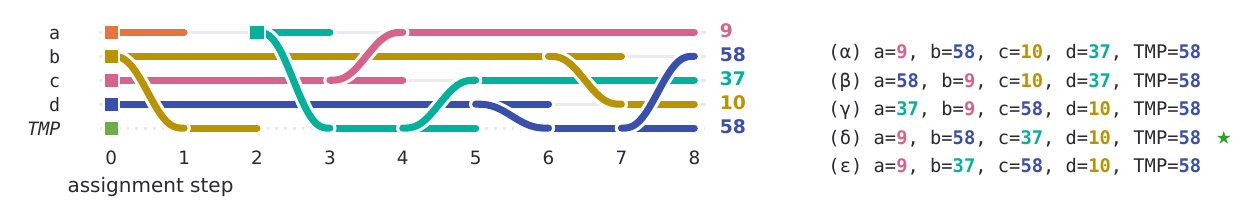}
    \vspace{-4ex}
    \caption{\textbf{left:} A visualization of an $8$-assignment general finite sequence with $4+1$ variables. Each variable is assigned an initial value (e.g., $\texttt{b} \gets 10$, $\texttt{c} \gets 9$). At step $t$, a variable may be overwritten by a new constant value (as in $\texttt{a} \gets 37$ at $t_2$) or it may be swapped with another (as \texttt{b} and \texttt{d} are swapped between $t_6$ and $t_8$). \textbf{right:} Five possible final states for the sequence, with the correct one ($\delta$) annotated with a green star.}
    \label{fig:state-bench-sample}
\end{figure}

To evaluate how well language models can learn and solve state-tracking problems we introduce StateBench, a new benchmark of synthetic state-tracking data that can be used to generate training and evaluation samples at targeted levels of algebraic complexity. Each StateBench problem is a monoid word problem over $m$ variables, initialized to starting values and updated with $n$ assignments. Assignments may be transpositions which swap two variables (finite), self-increment (soluble), reassignments by value (aperiodic), or constant assignments (r-trivial; see~\cref{fig:hierarchy}). Each sequence is paired with one correct and four incorrect candidate final states (\cref{fig:state-bench-sample}), supporting use as both a generative and a multiple-choice evaluation.

We first generate a general finite state tracking sample that mixes assignments-by-reference ($x \leftrightarrow y$, implemented with a temporary variable) with aperiodic assignments-by-value to a new constant value. For each sample, the number of aperiodic assignments is drawn log-uniformly from $[1,64]$. From this general finite state tracking source task, we derive a soluble sample by replacing all assignments-by-reference with the equivalent self-increment value ($x \gets x + \delta \bmod N$, for some max value $N$), an aperiodic sample by replacing it with the value it would have copied ($x \gets K$), and an r-trivial sample by replacing every assignment with the variable's initial value ($x \gets x_0$). Deriving all four from one source controls for the number of variables, the number of distinct values, and the positions of assignments across complexity levels.

We create a static evaluation set containing r-trivial, aperiodic, soluble, and finite samples with 8 to 64,000 assignments. Each sample is annotated with its complexity class, $m$, $n$, sensitivity, and length in Olmo-3 tokens; see~\Cref{tab:state-bench-stats} in \Cref{sec:extended-results}. Since pretrained models may be sensitive to surface form, each sample is rendered in six textual formats that share the same underlying algebraic structure (\Cref{sec:formattings}).

\section{Directly-Trained Model Evaluations}\label{sec:direct-training}

We first validate our theoretical placement of various language model architectures on the state-tracking hierarchy by training models directly on data from StateBench and evaluating their ability to generalize to longer sequences. We consider six architectures: Transformers \citep{vaswani-2017-attention} with rotary position embedding (RoPE; \citealt{su-2021-roformer}) and without any positional embedding (NoPE; \citealt{wang-2024-length}); Gated DeltaNet (GDN; \citealt{yang-2024-gated}) with only positive eigenvalues ($+$) and with both positive and negative eigenvalues ($\pm$); and Transformer-NoPE--GDN hybrid models with only positive eigenvalues ($+$) and both positive and negative eigenvalues ($\pm$) eigenvalues \citep{merrill-2026-olmo}. For each architecture we randomly initialize 275M-parameter models and train them with a language modeling objective on sequences of a StateBench prefix and its valid completion. We train each model on $1\times$ the compute-optimal amount of data for pretraining following the ``Chinchilla'' methodology of \citet{hoffmann-2022-training}. We consider two aspects of performance: first, whether models are able to \emph{learn} the task sufficiently well to attain near-ceiling performance on in-distribution numbers $n$ of assignments; and second, whether the models are able to \emph{generalize} to longer sequences with larger $n$. We train models with $n$ between 8 and 256, and evaluate them on sequences with $n$ up to 64,000 (i.e., 250$\times$ longer),
looking in particular at generalization up to $4\times$ the training length as well as all the way to 64,000.

\begin{figure}[ht]
    \centering
    \vspace{-2ex}
    \includegraphics[width=\linewidth]{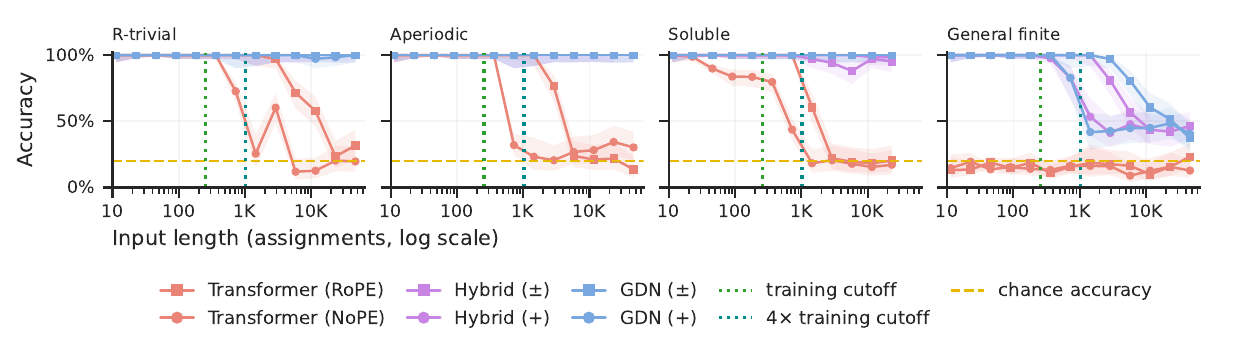}
    \vspace{-4ex}
    \caption{Direct Training: GDNs generalize perfectly on r-trivial, aperiodic, and soluble data, and partially on general finite data; Transformers fail to generalize beyond their training distribution on r-trivial, aperiodic, and soluble data and fail to even learn in-distribution general finite data. Error bars show Wilson confidence intervals.}
    \label{fig:dt-summary}
\end{figure}

\Cref{fig:dt-summary} shows that language model architectures show clear differences in their ability to learn state tracking at in-distribution lengths as well as to generalize to longer sequences. \textbf{Transformers} learn in-distribution \emph{r-trivial} and \emph{aperiodic} data well but do not generalize perfectly beyond the training distribution; NoPE models degrade immediately beyond the training cutoff, while RoPE models generalize to roughly $10\times$ the training length before degrading. On \emph{general finite} data, transformers do not even completely learn the in-distribution data. \textbf{GDN} models generalize perfectly on \emph{r-trivial}, \emph{aperiodic}, and \emph{soluble} data. They also perfectly learn the training distribution for \emph{general finite} data but fail to generalize perfectly beyond the training cutoff: GDN$^\pm$ models generalize to roughly $10\times$ the training lengths, whereas GDN$^+$ models degrade immediately beyond the training cutoff.
\textbf{Hybrid} models pattern like their GDN components, generalizing near-perfectly on r-trivial, aperiodic, and soluble data; Hybrid$^\pm$ models generalize longer than Hybrid$^+$ ones on general finite data.

\begin{figure}[ht]
    \centering
    \vspace{-2ex}
    \includegraphics[width=\linewidth]{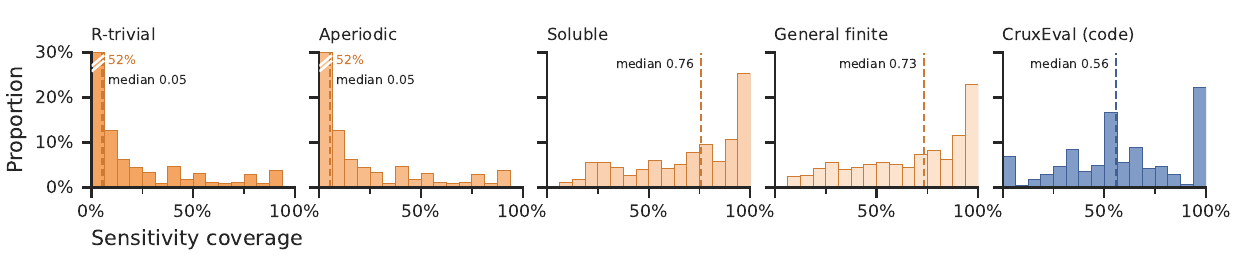}
    \vspace{-4ex}
    \caption{Sensitivity Analysis: Histograms of sample sensitivities in \textbf{(left)} the four complexity distributions in StateBench and \textbf{(right)} code samples from CruxEval.}
    \label{fig:sensitivity-histograms}
\end{figure}

\noindent \textbf{Sensitivity makes tracking state harder.} To interpret models' lackluster performance on general finite problems, we draw a connection between the periodicity of samples and their \emph{sensitivity}, a well-studied complexity measure that describes how much perturbations in a function's input affect its output.
Theoretically, monoid word problems have $o(n)$ sensitivity if they are aperiodic and $\Omega(n)$ otherwise \citep[Remark C.2]{amiri-2025-lower}. In other words, a state tracking problem is highly sensitive if and only if the underlying monoid is periodic.

We first measure the sensitivity of our derived evaluation samples and compare it to that of samples from real-world code evaluation data from CruxEval \citep{gu-2024-cruxeval}, a code-understanding benchmark of Python methods paired with canonical inputs and their expected outputs. In both cases, we measure sensitivity by calculating at how many steps during a program's execution modifying a variable's value could affect the final output.\footnote{For CruxEval samples, we take the codepath the function would take with the given input and unroll any loops; StateBench samples are already loop-free.} As~\cref{fig:sensitivity-histograms} shows, r-trivial and aperiodic data has low average sensitivity, while soluble and general finite sequences are far more sensitive, validating that they truly reflect periodic state tracking.
Code samples from CruxEval show a split distribution, with peaks at very high and very low sensitivities and a smoother distribution of sensitivities in the middle. Due to the correspondence between sensitivity and periodicity, this suggests periodic state tracking is important for evaluating natural code.

We next compare performance on general finite problems with per-sample sensitivity, finding a clear negative relationship: even among models that can learn general finite data in-distribution, samples with higher sensitivity are harder (cf.~\Cref{fig:sensitivity-acc} in \Cref{sec:extended-results}). This is aligned with prior observations that highly-sensitive functions are difficult for Transformers to learn \citep{hahn-2024-why}.
A data curriculum may aid learning sensitive state tracking problems \citep{siems-2026-learning,barak2022hidden,kearns98efficient}.
To test if this helps for our models, we compare the results of training on randomly-shuffled general finite sequences to the same training distribution ordered by ascending sensitivity. We find, however, that this has negligible effects on GDN and Hybrid models and on Transformers with NoPE (cf.~\Cref{fig:dt-sensitivity} in \Cref{sec:extended-results}), and only a weak positive effect on Transformers with RoPE on in-distribution general finite sequences.

\section{Pretrained Model Evaluations}

To measure how well language models solve state-tracking at differing levels of complexity without task-specific training, we also evaluate a suite of pretrained models on StateBench. We study two 7B parameter models: Olmo-3 \citep{teamOlmo-2025-olmo}, a Transformer-NoPE model; and Olmo-3-Hybrid \citep{merrill-2026-olmo}, a Hybrid$^\pm$ model, both pretrained, midtrained, and extended for long context on the Dolma dataset \citep{soldaini-2024-dolma}. We also study a family of smaller, custom pretrained models trained on the same dataset, at two size classes: 275M parameters (Transformer-NoPE, GDN$^\pm$, and Hybrid$^\pm$); and 2.7B parameters (Hybrid$^\pm$). Pretraining and midtraining data were capped at 8K tokens for all models, while long-context data was capped at 64K.

Since pre-trained models are frequently meant to be `task-agnostic' tools, evaluating on one particular formatting may bias results or make them unreflective of general state-tracking performance. To account for this, we generate six different sequence formats for all evaluation samples, including integer code, spreadsheet cell filling, and cube painting (cf.~\Cref{sec:formattings} for details and examples). We include sequences of lengths up to 128K to test how well models can perform on sequences with lengths seen during pretraining (first vertical green line in the plots), seen only during long-context extension (second vertical green line), and longer than have been seen at all during training.

\begin{figure}[ht]
    \centering
    \vspace{-2ex}
    \includegraphics[width=\linewidth]{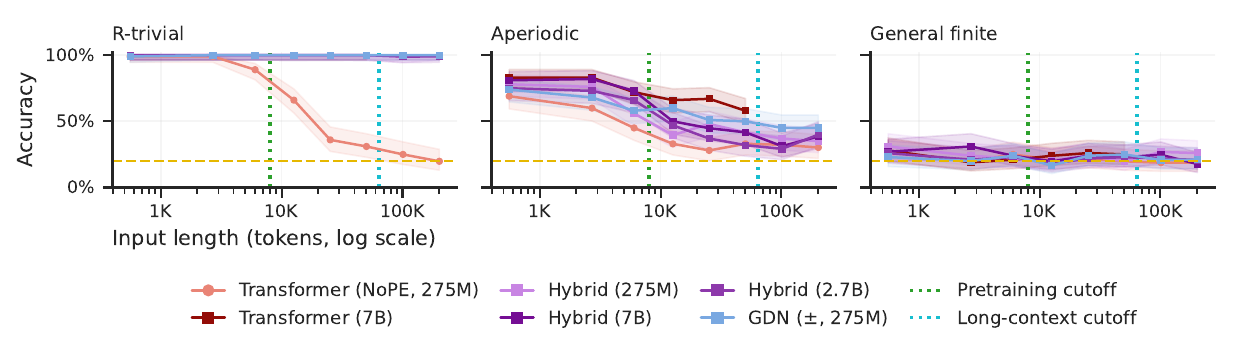}
    \vspace{-4ex}
    \caption{Pretrained models underperform theoretical expectations and direct-training results, showing consistent performance degradation on aperiodic data and completely failing on general finite data. Error bars show Wilson confidence intervals.}
    \label{fig:pt-results-max}
\end{figure}

\noindent \textbf{Pretrained models underperform at tracking state.} We evaluate models on all six different sequence formats, and report in \cref{fig:pt-results-max} the maximum value obtained by the model on any of those formats. Pretrained models consistently under-perform relative to theory and direct training. All models perform quite well on the r-trivial tasks except for the 275M-parameter transformer, which falls to at-chance accuracy beyond the pretraining length distribution. On aperiodic data, all models underperform expectations. The 7B models attain only $\sim$90\% accuracy, and all models exhibit some degree of performance degradation within the pretraining length window, and noticeably drop afterwards. Some models, particularly the 275M GDN$^\pm$ model, manage to maintain accuracy of $\sim$50\% beyond the long-context cutoff, but no higher. On general finite data, all models perform at-chance for all sequence lengths. In general, we find that \emph{Krohn--Rhodes complexity}, \emph{sample length}, and \emph{model size} are the dominant factors impacting accuracy.

\noindent \textbf{Models are sensitive to problem formatting.} We find models exhibit a strong preference for some task formats over others (cf.~\cref{fig:pt-per-task,fig:pt-results-by-format} in \Cref{sec:extended-results}). Data formatted like sequences from RULER \citep{hsieh-2024-ruler-a} are much easier for models than the exact same sequences formatted like manipulations of named spreadsheet cells. This is especially true for small models, which attain near-ceiling performance on RULER-formatted r-trivial data but are near-chance on spreadsheet-formatted data, but even 7B-parameter models also generalize better to longer sequences when they are formatted like RULER tasks. RULER-formatted samples also have lower aggregate variance. This suggests that pretrained models are strongly affected by the particular choice of formatting for state-tracking tasks, and hence that evaluations of state-tracking capabilities in pretrained models should take care to measure performance across a variety of different formats.

\begin{figure}[ht]
    \centering
    \vspace{-1ex}
    \includegraphics[width=\linewidth]{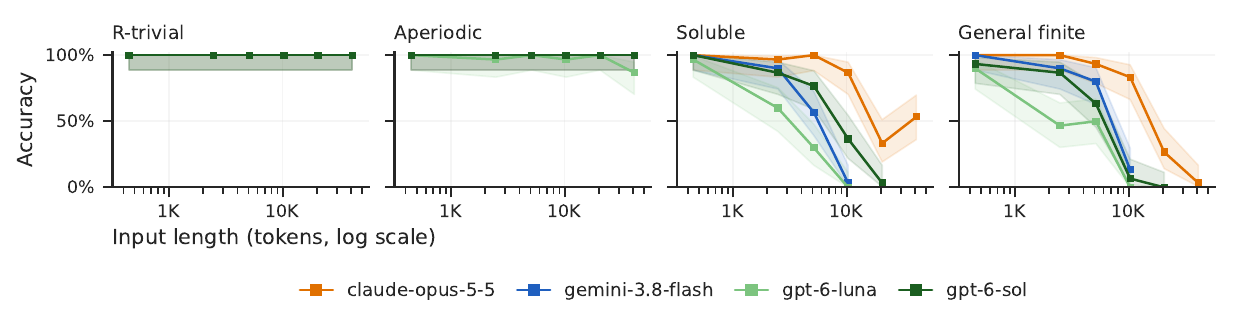}
    \vspace{-4ex}
    \caption{Frontier LLMs struggle on long-horizon soluble and general finite state-tracking tasks. Error bars show Wilson confidence intervals.}
    \label{fig:api-models}
\end{figure}

\noindent \textbf{Frontier models struggle on long-horizon hard state tracking.} We also run StateBench evaluations on four recently-released, proprietary frontier models: Gemini 3.8 Flash~\citep{google-2026-introducing}, Claude Opus 5.5~\citep{anthropic-2026-claude}, and GPT-6 Sol and Luna~\citep{openai-2026-introducing}. We evaluate these models in a generative setting, where they are asked to produce the final state of the system instead of choose among 5 possible answers. As shown in~\cref{fig:api-models}, all frontier models show strong performance on r-trivial and aperiodic problems; however, they fall to at-chance accuracy on soluble and finite state-tracking problems when inputs reach between 10K 64K tokens. The diminished performance of frontier reasoning LLMs shows that there is substantial headroom for improvement on generalizing on sensitive state-tracking problems.

\section{Conclusion}

We introduce a hierarchy of state tracking based on the Krohn--Rhodes decomposition of finite monoids, classifying problems as r-trivial, aperiodic, soluble, and general. We place various models of Transformers and Recurrent architectures within this hierarchy, showing the kinds of state-tracking tasks they can express length-general solutions to. We validate these results by constructing StateBench, a benchmark of state-tracking problems, and evaluate directly-trained, pretrained, and frontier language models on it. Our empirical findings support our theoretical results and highlighting the headroom left for language models to improve on state-tracking tasks.

One surprising takeaway from our empirical findings and novel theoretical results is that GDN with non-negative eigenvalues is capable of certain kinds of periodic state tracking, which is notable given prior work (cf.~\citealp{grazzi2025unlocking}).
Several questions remain open.
First, we do not know the exact placement of GDN$^+$
in the hierarchy. It would be interesting to understand better understand the relation of GDN$^+$ to soluble state tracking, as well as the impact of numerical precision or hybridization with attention layers.
Extending the hierarchy beyond finite monoids, i.e., to unboundedly-many variables, would bring many practical state-tracking tasks within the same framework, such as arbitrary discourse tracking. Empirically, our sensitivity analysis of code data suggests that general finite state tracking matters for code understanding, but where periodicity arises in natural data more broadly, and how much it drives the gaps between architectures in practice, remains an open question.

\subsection*{AI use statement}

The numerical constructions in \Cref{lem:gdn-beyond-aperiodic,lem:lrnn-beyond-aperiodic-exact} were developed interactively using Claude Opus 5.5 High and Gemini 3.6 Flash.
We used generative AI tools for code generation, including to help write the library used to generate state-tracking data and to help write plotting code used to generate figures. We also used generative AI tools to offer critiques on drafts of this paper and offer phrasing suggestions. We have not used generative AI to generate evaluation data directly, nor to directly generate any figures present in this paper.

We have reviewed all AI-assisted work through manual review. We take responsibility for the final content of this work, including text, claims, or artifacts produced with the aid of generative AI.

\subsection*{Reproducibility statement}

For all formal results, we include detailed definitions and proofs in~\cref{sec:formal-defs,sec:proofs}. For all empirical results, we provide a more detailed overview in~\cref{sec:formattings,sec:extended-results}. We will publicly release all data, code, and experimental setup conditions needed to reproduce our results in full.

\bibliography{iclr2027_conference,references}
\bibliographystyle{iclr2027_conference}

\appendix

\section{Formal Definitions}
\label{sec:formal-defs}

\subsection{Monoids and Krohn--Rhodes}

\begin{definition}[Finite Monoid]
    A \emph{monoid} is a tuple $(M,\circ,1)$ of a set $M$, an associative binary relation $\circ\colon M \times M \to M$, and an identity element $1$ such that $1 \circ a = a \circ 1 = a$ for all $a \in M.$ We say that a monoid is \emph{finite} if $M$ is finite.
\end{definition}

\begin{definition}[Word Problem]
    Let $M$ be a monoid, and let $w = w_1 \circ w_2 \circ \dots \circ w_k$ be a \emph{word} in $M$, i.e., a composition of finitely many elements $w_i \in M$. Since $M$ is closed under $\circ$, this composition is in $M$ as well. The \emph{word problem} on $M$ is to determine the identity of $w \in M$.
\end{definition}

\begin{definition}[Monoid Division]
    For monoids $A,B$, we say that $A$ \emph{divides} $B$, written $A \mid B$, if there exists a submonoid $B' \leq B$ and a surjective homomorphism $B' \twoheadrightarrow A$.
\end{definition}

\begin{definition}[Wreath Product] \label{defn:wreath-product}
    For finite monoids $M,N$ their wreath product $M \wr N$ is the set $M^N \times N$ of pairs $(f,n)$, where $f$ is a map from $N$ to $M$ and $n \in N$, under the composition
    \[
        (f, n)(g, n') \defeq (h, nn'),
        \qquad\text{where } h(x) \defeq f(x)\, g(xn) \text{ for all } x \in N,
    \]
    with identity $(\mathbf{1}, 1)$, where $\mathbf{1}(x) \defeq 1$ for all $x \in N$.
\end{definition}

Intuitively, $M \wr N$ is a \emph{cascade} in which $N$ processes an input and $M$ processes the input according to $N$'s current state.

\begin{definition}[Irreducibility]
    A finite monoid $M$ is irreducible if $M \mid (A \wr B)$ implies that $M \mid A$ or $M \mid B$ for all finite monoids $A,B$.
\end{definition}

\begin{theorem}[\citealt{krohn-1965-algebraic}]
    Let $M$ be a finite monoid. Then there exist finitely-many irreducible factors $M_1, \dotsc, M_k$ such that $M \mid (M_1 \wr M_2 \wr \dotsc \wr M_k)$ and each $M_i$ is either (a)~the flip-flop monoid $U_2 \defeq \{1,a,b\}$ where $xy = y$ for all $x,y \in \{a,b\}$; or~(b) a finite simple group that divides $M$.
\end{theorem}

\begin{definition}[R-trivial Monoid]
    A finite monoid $M$ is \emph{r-trivial} if $xM = yM$ implies $x = y$ for all $x,y \in M$.
\end{definition}

\begin{theorem}[\citealt{stiffler-1973-chapter}]
\label{thm:stiffler}
    A finite monoid $M$ is r-trivial if and only if $M \mid (U_1 \wr U_1 \wr \cdots \wr U_1)$ for some finite number of copies of $U_1$.
\end{theorem}

\begin{definition}[Aperiodic Monoid]
    A finite monoid $M$ is aperiodic if and only if $M \mid (U_2 \wr U_2 \wr \cdots \wr U_2)$ for some finite number of copies of $U_2$.
\end{definition}

\begin{definition}[Soluble Monoid]
    A finite monoid $M$ is soluble if and only if $M \mid (M_1 \wr M_2 \wr \cdots \wr M_k)$ for some finite $k$ and each $M_i$ is either (a) $U_2$ or (b) a cyclic group of prime order $\Z_p$ which itself divides $M$.
\end{definition}

\subsection{The Transformer Model}\label{app:transformer-model}

We adapt our formal model of transformers and RNNs from \citet{merrill-2026-why,merrill2025theory}.
Let $\Sigma$ be the token alphabet.
We first define our notion of a multi-layer sequence model that will consume a sequence $w \in \Sigma^*$ and produce a next token:

\begin{definition}[Multi-layer, multi-head sequence model; \citealt{merrill-2026-why,merrill2025theory}] \label{def:transformer}
A decoder-only transformer with $h$ heads, $d$ layers, model dimension $m$ (divisible by $h$), and feedforward width $w$ is parameterized by
\begin{enumerate}
    \item A fixed embedding matrix $\mathbf E: \mathbb Q^m \to \mathbb Q^{\Abs{\Sigma}} $;
    \item For $1 \leq \ell \leq d$ and $1 \leq k \leq h$, query, key, and value projections $\mathbf Q^\ell_k, \mathbf K^\ell_k, \mathbf V^\ell_k : \mathbb Q^m \to \mathbb Q^{m/h}$. We also specify for each $\ell, k$ the head type: SMAT, AHAT, or GDN.
    \item For $1 \leq \ell \leq d$, an output projection matrix $\mathbf O^\ell : \mathbb Q^m \to \mathbb Q^m$, initial feedforward projection $\mathbf W^\ell_1 : \mathbb Q^m \to \mathbb Q^w$, and final feedforward projection $\mathbf W^\ell_2 : \mathbb Q^w \to \mathbb Q^m$.
    \item An output projection matrix $\mathbf U : \mathbb Q^m \to \mathbb Q^{\Abs{\Sigma}}$.
\end{enumerate}
\end{definition}

We now define how such a model computes a function $\Sigma^* \to \Sigma$.

\begin{definition}[\citealt{merrill-2026-why,merrill2025theory}] \label{def:transformer-computation}
Given an input $w \in \Sigma^*$, we apply inductive rules for each component, producing internal representations and, finally, an output as follows:
\begin{enumerate}
    \item \underline{Embedding layer:} Let $\mathbf E[w_i]$ is the embedding of $w_i$. For $1 \leq i \leq n$, we define
    \begin{equation*}
        \mathbf h^0_i = \mathsf{layer\_norm}\left(\mathbf E[w_i]\right) .
    \end{equation*}
    \item \underline{Self-attention sublayer:} At each layer $\ell$, we compute $h$ attention heads.
    We first compute the sublayer input $\mathbf{\hat h}_i^{\ell-1} = \mathsf{layer\_norm}(\mathbf h_i^{\ell - 1})$.
    For each head $1 \leq k \leq h$, we compute queries $\mathbf q_{i,k}^\ell = \mathbf Q^\ell_k \mathbf {\hat h}^{\ell-1}_i$, keys $\mathbf k_{i,k}^\ell = \mathbf K^\ell_k \mathbf {\hat h}^{\ell-1}_i$, and values $\mathbf v_{i,k}^\ell = \mathbf V^\ell_k \mathbf {\hat h}^{\ell-1}_i$.
    The type of the head specifies a function that maps $\mathbf q_{i,k}^\ell, \mathbf k_{:,k}^\ell, \mathbf v_{:,k}^\ell$ to each head output $a^\ell_{i,k} \in \mathbb Q^{m/h}$ in a way we will return to defining.
    Finally, the self-attention sublayer then concatenates the output of $h$ heads and then applies a linear projection to compute
    \begin{equation*}
        \mathbf g^\ell_i = \mathbf h^{\ell - 1}_i + \mathsf{layer\_norm} \left( \langle \mathbf a^\ell_{i,1}, \ldots, \mathbf a^\ell_{i,h} \rangle \cdot \mathbf O^\ell \right) .
    \end{equation*}

    \item \underline{Feedforward sublayer:}
    Let $\mathbf{\hat g}_i^{\ell} = \mathsf{layer\_norm}(\mathbf g_i^{\ell})$.
    The feedforward sublayer computes the following tokenwise transformation:
    \begin{equation*}
        \mathbf h^{\ell}_i = \mathbf g^\ell_i + \mathbf W^\ell_2 (\mathbf W^\ell_1 \mathbf{\hat g}^\ell_i)_+
    \end{equation*}

    \item \underline{Unembedding layer:} The transformer outputs the token associated with $\arg \max \mathbf h^d_n \mathbf U$.
\end{enumerate}
\end{definition}

In this way, a multi-layer, multi-head sequence model $\mathcal M$ implements a function $\Sigma^* \to \Sigma$.
To view such models as language recognizers, we can simply identify a token in $\Sigma$ as a special ``accept'' symbol $a$ and say that the transformer recognizes a language $L$ if, for any string $w$, $w \in L \iff \mathcal M(w) = a$.

This architecture follows peri-norm \citep{kim2025periln}: the use of layer-norm both before and after each sublayer (as well as on the embeddings).
As discussed in \citet[Appendix A.1]{li2026efficientlyrepresenting}, this is capable of simulating projected pre-norm \citep{merrill-2023-expressive}, a standard tool in AHAT constructions.

\subsection{Sequence Mixers}

The last step is to define the types of sequence mixers used by each architecture:

\begin{definition}[SMAT Head]
    The output of head at token $i$ on queries $\mathbf q_i$, keys $\mathbf k_i$, and values $\mathbf v_i$ is
    \begin{equation*}
        \mathbf a_i = \sum_{j=1}^{i} \frac{\exp(1/\tau \cdot \mathbf q_i^\top \mathbf k_j )}{Z_i} \cdot \mathbf v_j ,
        \; \textrm{where} \;
        Z_i = \sum_{j=1}^{i} \exp(1/\tau \cdot \mathbf q_i^\top \mathbf k_j )
    \end{equation*}
    and $\tau = \sqrt{d/h}$.
\end{definition}

\begin{definition}[AHAT Head]
    The same as an SMAT head, but taking the limit $\tau \to 0$.
\end{definition}

\begin{definition}[GDN Head]
    The output of head at token $i$ is $\mathbf a_i$ defined via
    \[
        \mathbf S^\ell_i = \alpha^\ell_i \big(\mathbf I - \beta_i \mathbf k_i \mathbf k_i^\top\big) \mathbf S_{i-1} + \beta_i \mathbf k_i \mathbf v_i^\top,
        \qquad \mathbf a_i = \mathbf S_i^\top \mathbf q_i,
    \]
    with gate $\alpha_i \in [0,1]$ and write strength $\beta_i$ parameterized as in \citet{yang-2024-gated}.
    The eigenvalues of the transition $\mathbf I - \beta_i \mathbf k_i \mathbf k_i^\top$ are $1$ and $1 - \beta_i$. Following \citet{grazzi2025unlocking}, we write $\GDN^+$ when $\beta_i \in [0,1]$ (nonnegative eigenvalues) and $\GDN^\pm$ when $\beta_i \in [0,2]$ (eigenvalues in $[-1,1]$).
\end{definition}

By the classes $\SMAT, \AHAT, \GDN^+,$ and $\GDN^\pm$, we denote the classes of languages expressible when all heads are the corresponding type.

\subsection{Fixed-Precision Fixed-Point Overflow Arithmetic} \label{sec:fixed-precision}

We use the following fixed-precision arithmetic model in our analysis. Let $\D_{p,d}$ denote a datatype of numbers of the form $a/d$ where $d \in \Z_{>0}$ is a \emph{fixed} positive integer and $a \in (\Z \cap [-2^p, 2^p]) \cup \{\overflow\}$ is a bounded signed integer or a special overflow value. Each number in $\D_{p,d}$ can be specified using $p+2$ bits. When $d = 10^q$, $D_{p,d}$ corresponds to \textbf{fixed-point arithmetic} with $q$ decimal digits. Similarly, when $d = 2^q$, we have fixed-point arithmetic with $q$ fractional bits. $d$, however, does not have to take one of these forms; it can be any positive integer. We next discuss three important aspects of this datatype: overflow, left-to-right evaluation, and rounding.

When computing arithmetic operations over two numbers in $\D_{p,d}$, if the resulting value lies in the range $(1/d) [-2^p, 2^p]$, it is projected on to the nearest value (breaking ties upwards) in $\D_{p,d}$. If the resulting value would exceed $2^p / d$ or drop below $-2^p / d$, it goes to the \textbf{overflow state} $\overflow$ and cannot be further modified.

Expression over $\D_{p,d}$ are evaluated by applying arithmetic operators in a standard order. However, under this arithmetic model, operations are \emph{not associative}. For instance, $(1/d)(a + (2^p - 2^p))$ equals $a/d$, while for any $a > 0$, $(1/d)((a + 2^p) - 2^p)$ results in an overflow, yielding $\overflow$.\footnote{While some prior work has leveraged overflow in a controlled manner to induce desirable non-linear behavior in fixed-precision architectures, our constructions do not capitalize on overflow semantics.} To make the arithmetic well defined, we assume a \textbf{left-to-right evaluation order} for any iterated additions/subtractions as well as multiplications/divisions.

When two numbers $a/d$ and $b/d$ in $\D_{p,d}$ are added or subtracted without overflow, we obtain a result of the form $c/d$ that lies in $\D_{p,d}$. However, their product is $ab/d^2$, and when $d > 1$, this (again assuming no overflow) must be \textbf{rounded} out to the form $c/d$ where $c = [ab/d]$ denotes $ab/d$ rounded to the nearest integer, breaking ties upward. The larger the $d$ is, the stronger the rounding is. E.g., when $d = 10^q$, we effectively round away the $q$ least significant digits of the product. As we will see later (e.g., \cref{lem:gdn-beyond-aperiodic}), this rounding behavior plays an important role in simulating certain state tracking primitives with GDNs.

\noindent\emph{Remark.}
Going beyond fixed precision would break the upper bounds we report.
Further, even within fixed-precision, as discussed by \citet{nowak-2026-algebraic}, the expressivity of linear RNNs can be sensitive to the choices in the specific arithmetic model used. For example, if large values \emph{wrapped around} to 0 rather than overflowing, this would allow simulating $\Z_p$, adding expressive power to otherwise aperiodic architectures.
Finally, we leave it to future work to understand how different evaluation orders might affect expressivity.

\section{Expressivity Proofs}
\label{sec:proofs}

\subsection{Basic Monoids}

\basicMonids*

\begin{proof}[Proof of \Cref{lem:building-blocks}.\ref{lem:smat_U2}]
    To recognize word problems in the monoid $U_1 = \{1, 0\}$, we simply need to check if any $0$ occurs in the context $w$.
    We do this by constructing an attention head $h$ where the query is $1$, the key $j$ a boolean encoding whether token $w_j = 0$, and the value is $1$.
    $h$ will return $1$ iff there exists some $j$ such that $w_j = 0$.
    Thus, taking $1 - h$ will compute the word problem for $U_1$.
\end{proof}

\begin{proof}[Proof of \Cref{lem:building-blocks}.\ref{lem:gdn-flip-flop}]
    Ignoring the gating, GDN$^+_f$ has the form
    \begin{equation*}
        \mathbf S_t = (\mathbf I - \beta_t \mathbf k_t \mathbf k_t^\top) \mathbf S_{t-1} + \changed{\beta_t} \mathbf k_t \mathbf v_t^\top .
    \end{equation*}
    Fixing a vector dimension $d$, let $\vec{i}$ be the one-hot vector for $i \leq d$.
    If $\mathbf k_t = \vec{i}$, $\mathbf v_t = \vec{j}$, and $\beta_t = 1$, then the GDN update can be understood to replace the vector in row $i$ of the state matrix with $\vec j$.

    We now describe the updates required to implement $U_2$.
    For $w_t \in \{a, b\}$, let $\mathbf k_t = \vec 1$ and $\mathbf v_t$ be $\vec 1$ or $\vec 2$, respectively, with $\beta_t = 1$ in either case.
    Thus, these updates overwrite the first row of $\mathbf S_t$ to be either $\vec 1$ or $\vec 2$.
    For $w_t = 1$, let $\beta_t = 0$, reducing the update to the identity.
    Thus, the first row of $\mathbf S_t$ is a representation of the $U_2$ word problem output, with $\vec 0 \equiv 1$, $\vec 1 \equiv a$ and $\vec 2 \equiv b$.
\end{proof}

\begin{proof}[Proof of \Cref{lem:building-blocks}.\ref{lem:ahat-flip-flop}]
    To compute the $U_2$ word problem, we construct an AHAT attention head that attends over all tokens $a, b$ in the context and retrieves the rightmost one.
    To do this, we add two terms to key $j$: the first term is a boolean indicator for whether $w_j \in \{a, b\}$, and the second is tie-breaking term in $(0, 1)$ that monotonically increases with $j$ \citep[Appendix E]{merrill-2023-expressive}.
    The query at $i$ is just $1$, and the value at $j$ is simply $w_j$.
    Thus, at token $i$, the AHAT head returns the rightmost $w_j$ such that $w_j \in \{a, b\}$, effectively solving the word problem.
\end{proof}

\begin{proof}[Proof of \Cref{lem:building-blocks}.\ref{lem:ahat-mod}]
    Let representations of $z_1, z_2, \ldots z_n \in \{0, 1, \ldots, p-1\}$ be the input to a log-precision AHAT layer. To implement $\Z_p$, we would like it to compute $S_i \defeq \big(\sum_{j=1}^i z_j\big) \bmod p$ at each token $i$. The ``division lemma'' of \citet{merrill-2025-depth} is a valuable tool here: it shows how to compute $a \bmod p$ from a representation of $a$ at position $i$ using a few layers of AHAT. However, it assumes $a \leq i$, whereas our partial sum can be as large as $(p-1) \, i$. We get around this hurdle by leveraging the fact that $p$ is a fixed constant, as follows.
    
    First, for each $r \in \{0, 1, \ldots, p-1\}$, we construct a log-precision uniform attention head $h_r$ that computes the representation of $c_r \defeq |\{j \leq i \mid z_j = r\}|$, i.e., the \emph{number} of input values $z_j$ for $j \leq i$ that equal $r$. By definition, $c_r \leq i$. We can rewrite $S_i$ in terms of $c_r$ as follows:
    \begin{align*}
        S_i & = \left(\sum_{r=1}^{p-1} r \, c_r\right) \bmod p \\
            & = \left(\sum_{r=1}^{p-1} r \, (c_r \bmod p) \right) \bmod p
    \end{align*}
    Since $c_r \leq i$, we can now apply the division lemma of \citet{merrill-2025-depth} to the output of $h_r$ in order to compute $c_r \bmod p$ for each $r$ in parallel, using a total of $7$ AHAT layers. An MLP layer can then multiply the result (which is at most $p-1$) with $r$, resulting in a representation of $b_r \defeq r \, (c_r \bmod p)$, which is at most $(p-1)^2$. Another MLP layer can now compute $\sum_{r=1}^{p-1} b_r$, which is at most $(p-1)^3$. A final MLP layer can now apply a $\bmod p$ operation to this sum, resulting in $S_i$. We thus have an 8-layer log-precision AHAT that implements $Z_p$.
\end{proof}

\monoidComposition*

\begin{proof}
    We prove this by induction over the number of composed monoids.
    In the base case of $k = 0$, we simply have the input word $w$ (equivalently, the identity applied to $w$) encoded in the initial layer.
    For the inductive step, assume layer $\sum_{i=1}^{k-1} d_i$ has a representation of the partial products for input $w$ for the monoid $M_1 \wr M_2 \wr \cdots \wr M_{k-1}$.
    We have a monoid $M_k$ whose elements are from $f_k(M_1 \wr M_2 \wr \cdots \wr M_{k-1})$, where $f_k$ is a map from $M_1 \wr M_2 \wr \cdots \wr M_{k-1}$ to $M_k$ (cf.~\Cref{defn:wreath-product}).
    We can first compute a representation of these elements using a feedforward network since $f_k$ is simply a finite lookup table of the already represented partial products.
    By assumption, network $A_k$ can implement $M_k$ with a block of $d_k$ layers.
    We augment our inductively composed network \smash{$\tilde{A} = A_{k-1} \circ \ldots \circ A_2 \circ A_1$} with $d_k$ layers of $A_k$. In case $A_k$ operates over a different datatype than $\tilde{A}$, we can first cast the output representations of $\tilde{A}$ to the corresponding nearest values in the datatype of $A_k$. Layer $\sum_{i=1}^k d_i$ of this composed network will thus have a representation of the partial products for input $w$ under $M_1 \wr M_2 \wr \cdots \wr M_k$, finishing the inductive step.
\end{proof}

\subsection{Linear RNNs with Non-Negative Eigenvalues}

\gdnAperiodic*

We divide the proof of \Cref{thm:gdn-equals-aperiodic} into \Cref{lem:aperiodic-gdn,lem:gdn-beyond-aperiodic}.
We first show that GDN with positive eigenvalues can recognize any aperiodic language via the Krohn-Rhodes decomposition. This resembles Theorem 4 of \citet{sarrof-2024-expressive}, but the construction is instantiated for GDN$^+$ rather than an arbitrary linear RNN, and the arithmetic model assumed is slightly different.

\begin{lemma} \label{lem:aperiodic-gdn}
    Let $M$ be an aperiodic monoid. Then there exists a fixed-precision GDN$^+$ that solves the word problem over $M$.
\end{lemma}

\begin{proof}
    By \Cref{def:monoid-types}, $M$ has a KR decomposition $M_1 \wr M_2 \wr \ldots \wr M_k$ such that every $M_i$ is the flip-flop monoid $U_2$. By \Cref{lem:building-blocks}.\ref{lem:gdn-flip-flop}, $U_2$ can be implemented with a single GDN$^+_f$ layer. Applying \Cref{lem:composition}, we obtain a $k$-layer GDN$^+$ that implements $M$.
\end{proof}

\gdnBeyondAperiodic*

\begin{proof}
    On input $a$, define the key $k = \frac{1}{5} \langle -3, 4 \rangle$, which satisfies $\Abs{k} = 1$. It follows that
    \begin{equation*}
        A = I - kk^\top = \frac{1}{25}
        \begin{pmatrix}
            16 & 12 \\
            12 & 9
        \end{pmatrix} .
    \end{equation*}
    It suffices to show that there exist two states $h, h'$ such that $Ah = h'$ and vice versa:
    we can use $h = \frac{1}{25} \langle 9, 8 \rangle$ and $h' = \frac{1}{25} \langle 10, 7 \rangle$.
    Letting $[x]$ denote $x$ rounded to the nearest integer (breaking ties upward), we have
    \begin{align*}
        A h 
        &= 
        A \, \cdot \, \frac{1}{25}
        \begin{pmatrix}
            9 \\
            8
        \end{pmatrix}
        = \frac{1}{25} \begin{pmatrix}
            \left[\frac{16\times9}{25}\right] + \left[\frac{12\times8}{25}\right] \\[6pt]  %
            \left[\frac{12\times9}{25}\right] + \left[\frac{9\times8}{25}\right]
        \end{pmatrix}
        = \frac{1}{25} \begin{pmatrix}
            6 + 4 \\
            4 + 3
        \end{pmatrix}
        = \frac{1}{25} \begin{pmatrix}
            10 \\
            7
        \end{pmatrix}
        = h' \\[6pt]  %
        A h'
        &= A \, \cdot \, \frac{1}{25}
        \begin{pmatrix}
            10 \\
            7
        \end{pmatrix}
        = \frac{1}{25} \begin{pmatrix}
            \left[\frac{16\times10}{25}\right] + \left[\frac{12\times7}{25}\right] \\[6pt]  %
            \left[\frac{12\times10}{25}\right] + \left[\frac{9\times7}{25}\right]
        \end{pmatrix}
        = \frac{1}{25} \begin{pmatrix}
            6 + 3 \\
            5 + 3
        \end{pmatrix}
        = \frac{1}{25} \begin{pmatrix}
            9 \\
            8
        \end{pmatrix}
        = h .
    \end{align*}
    To leverage this to recognize $(aa)^*$, we use use the value associated with the beginning-of-sequence token to initialize the GDN state to $h$.
    Then we can detect which state we are in using a linear projection that checks whether $h_1 - h_2 > 2$.
    For the product $A^k$, this detects whether $k$ is even or odd, allowing us to recognize $(aa)^*$.

    This construction works for any datatype $\D_{p,25}$ where $p$ is large enough to represent the constants needed to perform the above computation.
\end{proof}

The above construction uses a fixed-precision fixed-point datatype with denominator $d = 25$. Such a construction is also possible when $d$ is a power of $10$, for instance when $d = 100$, i.e., rounding to two decimal places, which is perhaps more natural. In this case, we can set $k = \langle -0.8, 0.6 \rangle$, which has unit norm and yields $A = \left(\begin{smallmatrix} 0.36 & 0.48 \\ 0.48 & 0.64 \end{smallmatrix}\right)$. Let $h = \langle 0.12, 0.15 \rangle$ and $h' = \langle 0.11, 0.16 \rangle$. As seen below, $Ah = h'$ and $Ah' = h$ under rounding to two decimal places, denote $[x]_2$:
    \begin{align*}
        A h 
        &= 
        A
        \begin{pmatrix}
            0.12 \\
            0.15
        \end{pmatrix}
        = \begin{pmatrix}
            \left[0.36\times0.12\right]_2 + \left[0.48\times0.15\right]_2 \\[6pt]  %
            \left[0.48\times0.12\right]_2 + \left[0.48\times0.15\right]_2
        \end{pmatrix}
        = \begin{pmatrix}
            0.04 + 0.07 \\
            0.06 + 0.10
        \end{pmatrix}
        = \begin{pmatrix}
            0.11 \\
            0.16
        \end{pmatrix}
        = h' \\[6pt]  %
        A h'
        &= A 
        \begin{pmatrix}
            0.11 \\
            0.16
        \end{pmatrix}
        = \begin{pmatrix}
            \left[0.36\times0.11\right]_2 + \left[0.48\times0.16\right]_2 \\[6pt]  %
            \left[0.48\times0.11\right]_2 + \left[0.48\times0.16\right]_2
        \end{pmatrix}
        = \begin{pmatrix}
            0.04 + 0.08 \\
            0.05 + 0.10
        \end{pmatrix}
        = \begin{pmatrix}
            0.12 \\
            0.15
        \end{pmatrix}
        = h .
    \end{align*}
This provides another construction, with $d = 100$, supporting \Cref{lem:gdn-beyond-aperiodic}.

Further, as we show next, rounding isn't necessary if we don't limit ourselves to GDN style RNNs. Specifically, general linear RNNs with non-negative eigenvalues can recognize certain periodic languages even under \emph{exact} arithmetic:

\begin{lemma} \label{lem:lrnn-beyond-aperiodic-exact}
There exists a linear RNN where $A_t$ has only non-negative eigenvalues that recognizes $(abab)^*$ under datatype $D_{p,1}$ for all $p > 0$.
\end{lemma}

\begin{proof}
    Define the linear RNN's transition matrices on inputs $a$ and $b$ as
    $$A(a) =
    \begin{pmatrix}
        1&1 \\
        0&1
    \end{pmatrix} ,\qquad A(b) =
    \begin{pmatrix}
        1 & 0 \\
        -1&1
    \end{pmatrix}.$$
    Note that these transition matrices are not parameterizable as GDNs, and instead form general linear RNNs. Their product is
    \begin{equation}\label{eqn:parity-transition-matrix}
        A = A(b)A(a) = \begin{pmatrix}1&1\\-1&0\end{pmatrix}
    \end{equation}
    It can be verified that $A$ corresponds to a rotation with eigenvalues $e^{\pm i \pi / 3}$, which implies that $A^6 = I$.
    Assume the beginning-of-sequence token creates the initial state as a diagonal matrix with diagonal entries $1$ and $0$, represented here as $\langle 1, 0 \rangle$. Then, upon seeing input $(a b)^*$, we will cycle through the following 6 states at every $b$:
    \begin{align*}
        h_0 &= \langle 1, 0 \rangle \\
        h_1 &= \langle 1, -1 \rangle \\
        h_2 &= \langle 0, -1 \rangle \\
        h_3 &= \langle -1, 0 \rangle \\
        h_4 &= \langle -1, 1 \rangle \\
        h_5 &= \langle 0, 1 \rangle .
    \end{align*}
    Note that $h_3 = -h_0$ and $h_5$ in the next step cycles back to $h_0$.
    Each of these states can be detected via a simple linear projection.
    On input of the form $(ab)^k$, this yields a one-hot representation of $k \bmod 6$ in the residual stream.
    Applying a finite lookup table allows us to determine $k \bmod 2$ and recognize $(abab)^*$.
\end{proof}

The construction in the above proof, in fact, shows that if we allow $A_t$ to be parameterized as the product of matrices with non-negative eigenvalues, then this idea extends to even recognizing the parity language $(aa)^*$. Specifically, following \cref{eqn:parity-transition-matrix}, we can set
\begin{equation}
    A(a)
    = \begin{pmatrix}
        1 & 1 \\
        0 & 1
      \end{pmatrix}
      \begin{pmatrix}
        1 & 0 \\
        -1 & 1
      \end{pmatrix}
    = \begin{pmatrix}
        1 & 1 \\
        -1 & 0
      \end{pmatrix}
\end{equation}
where the two constituent matrices have only non-negative eigenvalues.
This product structure is similar to that of DeltaProduct~\citep{siems-2025-deltaproduct}, which parameterizes $A_t$ as the product of DeltaNet matrices. We thus have the following:

\begin{corollary}\label{cor:product-lrnn-aa-star}
    There exists a linear RNN where $A_t$ is parameterized as the product of matrices with only non-negative eigenvalues that recognizes $(aa)^*$ under datatype $\mathbb D_{p,1}$ for all $p > 0$.
\end{corollary}

\subsection{Log-Precision Attention}

\ahatSoluble*

\begin{proof}
    By \Cref{def:monoid-types}, $M$ has a KR decomposition $M_1 \wr M_2 \wr \ldots \wr M_k$ such that every $M_i$ is either the flip-flop monoid $U_2$ or $\Z_p$ for some prime $p$. By \Cref{lem:building-blocks}.\ref{lem:ahat-flip-flop}, $U_2$ can be implemented with a single AHAT$_\ell$ layer. By \Cref{lem:building-blocks}.\ref{lem:ahat-mod}, $Z_p$ can be implemented by a $8$-layer AHAT$_\ell$. Applying \Cref{lem:composition}, we obtain that AHAT$_\ell$ can implement $M$. More specifically, if $M_1 \wr M_2 \wr \ldots \wr M_k$ contains $r \leq k$ groups of the form $\Z_p$, then there exists a $(k + 7r)$-layer AHAT$_\ell$ that implements $M$.
\end{proof}

\begin{proposition} \label{prop:conditional-hybrid-soluble}
    Hybrid $(\GDN^+_f : \SMAT_\ell : \modgadget) \supseteq \Sol$, i.e., for any soluble $M$, there exists a hybrid model composed of fixed-precision \smash{GDN$^+$}, log-precision SMAT, and mod gadgets \smash{$\modgadget^p$} for a fixed set of value $p$ that solves the word problem over $M$.
\end{proposition}

\begin{proof}
    By \Cref{def:monoid-types}, $M$ has a KR decomposition $M_1 \wr M_2 \wr \ldots \wr M_k$ such that every $M_i$ is either the flip-flop monoid $U_2$ or the group $\Z_p$ for some prime $p$. By \Cref{lem:building-blocks}.\ref{lem:gdn-flip-flop}, $U_2$ can be implemented with a single fixed-precision GDN$^+$ layer.
    What remains to implement is $\Z_p$. Let representations of $z_1, z_2, \ldots, z_n \in \{0, 1, \ldots, p-1\}$ be the input to a single layer of log-precision SMAT. At token $i$, this layer can use uniform attention to compute a representation of the sum $y_i \defeq \sum_{j=1}^i z_j$, e.g., $1 / y_i$ or $\phi(y_i)$ or $y_i / i$.
    We can now feed this to the mod gadget $\modgadget^p$ to compute a representation of $z \bmod p$, thereby implementing $\Z_p$.
    Applying \Cref{lem:composition} shows that the full hybrid model can implement $M$.
\end{proof}
\section{Formatting Examples}
\label{sec:formattings}

Below are instances of the abstract general finite sample from~\cref{fig:state-bench-sample} rendered in the six different formats we use.

\subsection{Integer Code}
\begin{lstlisting}[style=sbprose]
Track the current value of each variable after executing the assignments.

// ........
a <- 8;
c <- 10;
// ........
// ........
TMP <- 81;
b <- 11;
// ........
d <- 59;
TMP <- b;
// ........
a <- 38;
TMP <- a;
a <- c;
c <- TMP;
TMP <- d;
d <- b;
b <- TMP;

a ?= 9; b ?= 58; c ?= 37; d ?= 10; TMP ?= 58;
\end{lstlisting}

\subsection{Spreadsheet Cells}
\begin{lstlisting}[style=sbprose]
Track the contents of each spreadsheet cell after following the instructions.

Enter 8 in cell A1.
......................................
......................................
......................................
Enter 10 in cell C1.
......................................
Enter 81 in scratch cell.
Enter 11 in cell B1.
Enter 59 in cell D1.
Copy the current contents of cell B1 into scratch cell.
Enter 38 in cell A1.
Copy the current contents of cell A1 into scratch cell.
Copy the current contents of cell C1 into cell A1.
......................................
Copy the current contents of scratch cell into cell C1.
Copy the current contents of cell D1 into scratch cell.
Copy the current contents of cell B1 into cell D1.
Copy the current contents of scratch cell into cell B1.

Final cell contents:

cell A1 contains 10. cell B1 contains 59. cell C1 contains 38. cell D1 contains 11. scratch cell contains 59.
\end{lstlisting}

\subsection{Cube Painting}
\begin{lstlisting}[style=sbprose]
Determine the final color of every cube after following the instructions.

Paint cube 1 teal.
...............................
Paint cube 3 gray.
Paint mixing cube topaz.
Paint cube 2 maroon.
Paint cube 4 brick.
...............................
Paint mixing cube the same color as cube 2.
Paint cube 1 ivory.
Paint mixing cube the same color as cube 1.
Paint cube 1 the same color as cube 3.
Paint cube 3 the same color as mixing cube.
Paint mixing cube the same color as cube 4.
Paint cube 4 the same color as cube 2.
Paint cube 2 the same color as mixing cube.
...............................
...............................
...............................

Final colors:

cube 1 is gray. cube 2 is brick. cube 3 is ivory. cube 4 is maroon. mixing cube is brick.
\end{lstlisting}

\subsection{People in Rooms}
\begin{lstlisting}[style=sbprose]
Track where each person is after following the movements in order.

Alice enters the basement.
.................................
Carol enters the studio.
.................................
.................................
Taylor enters the toolshed.
Bob enters the lounge.
David enters the lighthouse.
Taylor moves to Bob's current location.
Alice enters the canteen.
Taylor moves to Alice's current location.
Alice moves to Carol's current location.
Carol moves to Taylor's current location.
Taylor moves to David's current location.
David moves to Bob's current location.
.................................
Bob moves to Taylor's current location.
.................................

Final locations:

Alice is in the studio. Bob is in the lighthouse. Carol is in the canteen. David is in the lounge. Taylor is in the lighthouse.
\end{lstlisting}

\subsection{Status Lights}
\begin{lstlisting}[style=sbprose]
Track the color displayed by each status light after following the instructions.

Set light A to teal.
Set light C to gray.
Set buffer light to topaz.
Set light B to maroon.
..........................................
Set light D to brick.
Set buffer light to the color currently displayed by light B.
Set light A to ivory.
Set buffer light to the color currently displayed by light A.
Set light A to the color currently displayed by light C.
..........................................
Set light C to the color currently displayed by buffer light.
..........................................
Set buffer light to the color currently displayed by light D.
..........................................
Set light D to the color currently displayed by light B.
Set light B to the color currently displayed by buffer light.
..........................................

Final light colors:

light A displays gray. light B displays brick. light C displays ivory. light D displays maroon. buffer light displays brick.
\end{lstlisting}

\subsection{RULER}
\begin{lstlisting}[style=sbprose]
Some special magic numbers are hidden within the following text. Make sure to memorize it. I will quiz you about the numbers afterward.

One of the special magic numbers for hollow is: 9931280.
must have been when they first started to. It was presumably many
One of the special magic numbers for golden is: 3017169.
thousands of years between when people first started describing things as
One of the special magic numbers for temporary is: 8434201.
One of the special magic numbers for silver is: 5570834.
One of the special magic numbers for crimson is: 1829661.
The special magic number for temporary becomes the number currently assigned to silver.
hot or cold and when someone asked ''what is heat?'' No doubt it was a very
One of the special magic numbers for hollow is: 9580157.
The special magic number for temporary becomes the number currently assigned to hollow.
gradual process. We don't know if Plato or Aristotle were the first to
ask any of the questions they did. But their works are the oldest we have
The special magic number for hollow becomes the number currently assigned to golden.
The special magic number for golden becomes the number currently assigned to temporary.
The special magic number for temporary becomes the number currently assigned to crimson.
The special magic number for crimson becomes the number currently assigned to silver.
The special magic number for silver becomes the number currently assigned to temporary.

hollow is 3017169. silver is 1829661. golden is 9580157. crimson is 5570834. temporary is 1829661.
\end{lstlisting}

\section{Additional Statistics and Results}
\label{sec:extended-results}

\begin{table}[ht]
\centering
\begin{tabularx}{0.9\linewidth}{@{}X *{4}{S[table-format=2.2]}@{}}
\toprule
 & \multicolumn{4}{c}{Monoid Complexity Class} \\
\cmidrule(l){2-5}
 & {R-trivial} & {Aperiodic} & {Soluble} & {General finite} \\
\midrule
Max.\ dependency length (K) & 0    & 0    & 21.3 & 32.1 \\
\addlinespace
Sensitivity                 & 0.08 & 0.08 & 0.76 & 0.75 \\
                            & {\rng{0.00}{0.91}} & {\rng{0.00}{0.91}} & {\rng{0.06}{1.00}} & {\rng{0.04}{1.00}} \\
\addlinespace
Prompt tokens (K)           & 1.16 & 1.16 & 1.34 & 1.25 \\
                            & {\rng{0.07}{981}} & {\rng{0.07}{981}} & {\rng{0.09}{768}} & {\rng{0.06}{939}} \\
\midrule
\multicolumn{5}{@{}l}{\emph{Shared across classes}} \\
\addlinespace[2pt]
\quad Samples per condition & \multicolumn{4}{c}{\num{2720}} \\
\quad Variables ($m$)       & \multicolumn{4}{c}{2, 4, 8, 16} \\
\quad Assignments ($n$)     & \multicolumn{4}{c}{49 \rng{11}{64K}} \\
\bottomrule
\end{tabularx}
\caption{Descriptive statistics for StateBench. Cells give medians, with [min, max] ranges in grey below. Token counts and dependency lengths are in thousands.}
\label{tab:state-bench-stats}
\end{table}

\begin{figure}[ht]
    \centering
    \vspace{-3ex}
    \includegraphics[width=0.9\linewidth]{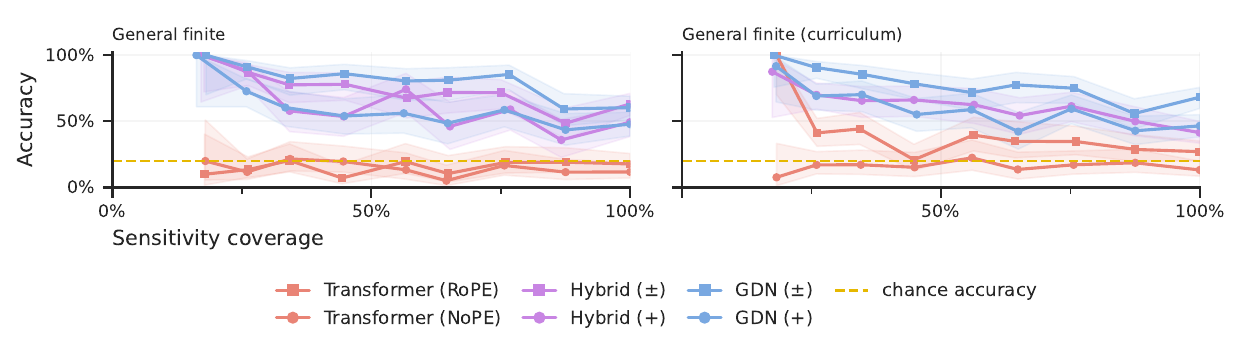}
    \caption{Effect of Sensitivity: On general finite state tracking data, direct trained models perform worse on higher sensitivity data. Error bars show Wilson confidence intervals.}
    \label{fig:sensitivity-acc}
\end{figure}

\begin{figure}[ht]
    \centering
    \includegraphics[width=0.9\linewidth]{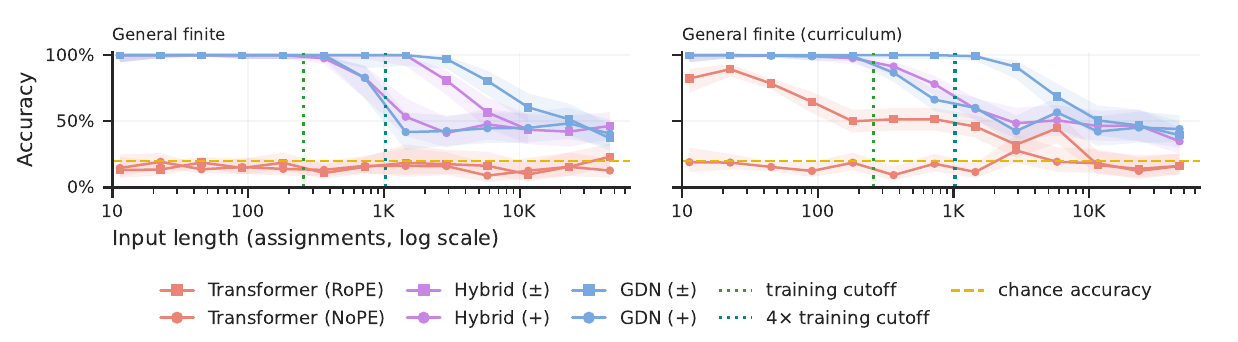}
    \caption{Curriculum Training: Sorting general finite sequences by sensitivity during training \textbf{(right)} boosts direct training performance of Transformer-RoPE models on in-distribution data compared to training on the same randomly-shuffled sequences \textbf{(left, repeated)}; Transformer-NoPE, GDN, and Hybrid models are unaffected. Error bars show Wilson confidence intervals.}
    \label{fig:dt-sensitivity}
\end{figure}

\begin{figure}[ht]
    \centering
    \includegraphics[width=\linewidth]{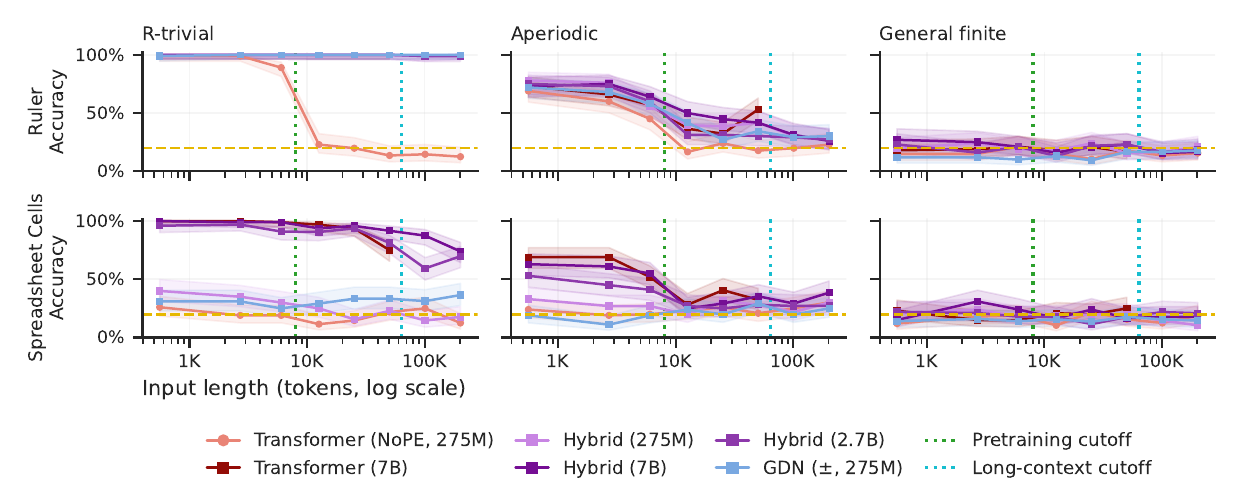}
    \caption{Pretrained base models perform better on RULER-formatted StateBench sequences \textbf{(top)} than on the same sequences formatted as spreadsheet manipulations \textbf{(bottom)}. Error bars show Wilson confidence intervals.}
    \label{fig:pt-per-task}
\end{figure}

\begin{figure}[ht]
    \centering
    \includegraphics[width=\linewidth]{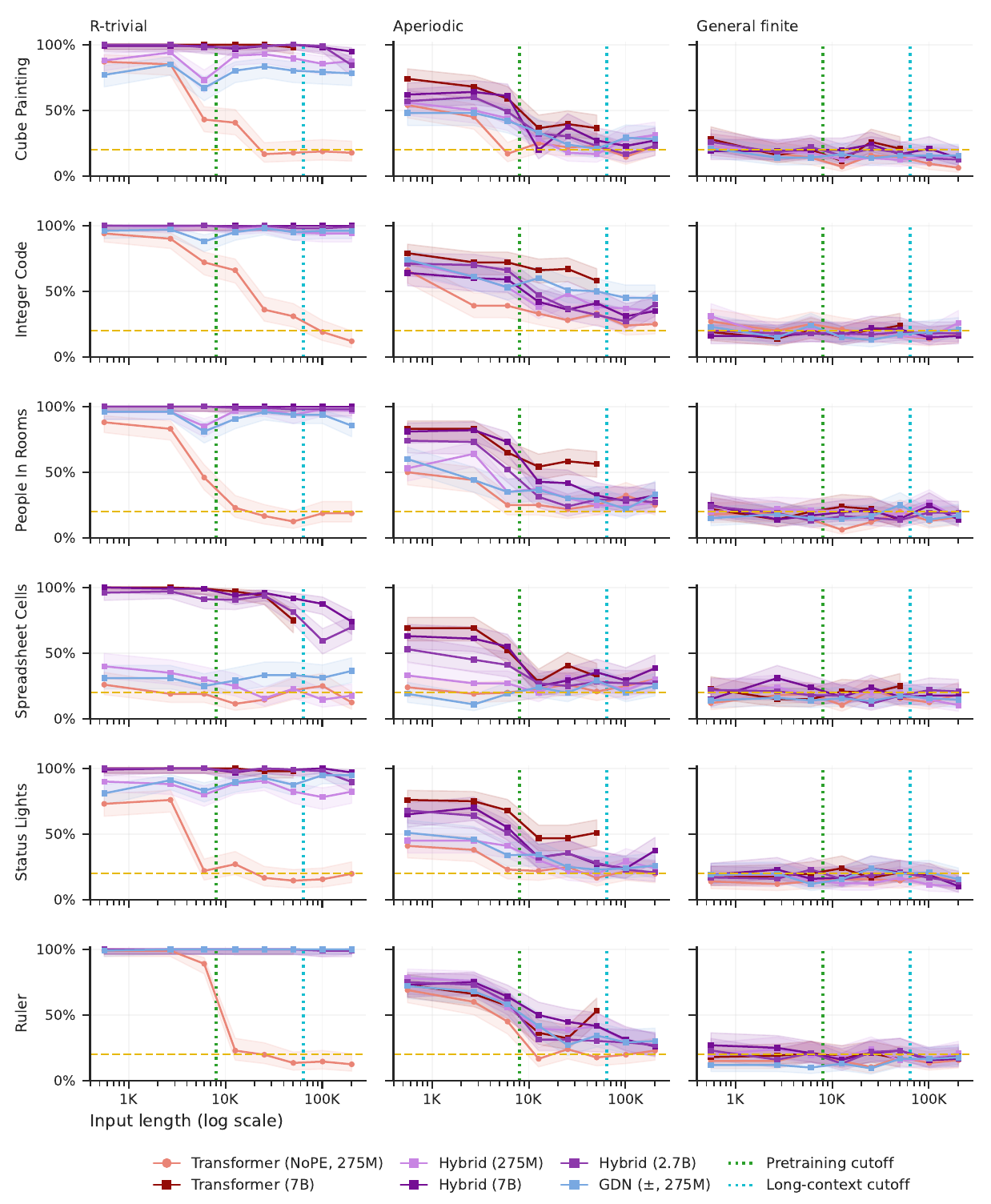}
    \caption{Pretrained models evaluated on StateBench problems by complexity class and formatting. Error bars show Wilson confidence intervals.}
    \label{fig:pt-results-by-format}
\end{figure}

\end{document}